\documentclass[10pt,journal,compsoc]{IEEEtran}
\usepackage{booktabs} 
\usepackage{multirow}
\usepackage{graphicx}
\usepackage{subfigure}
\usepackage{enumitem}
\usepackage{multirow}
\usepackage{amssymb}
\usepackage{amsmath}
\usepackage{xurl}
\usepackage{bbm} 
\usepackage{xcolor}

\ifCLASSOPTIONcompsoc
  \usepackage[nocompress]{cite}
\else
  \usepackage{cite}
\fi
\ifCLASSINFOpdf
\else
\fi
\begin{document}
%
\title{STRelay: A Universal Spatio-Temporal Relaying Framework for Location Prediction over Human Trajectory Data}
%
%
%
%

\author{Bangchao~Deng,~
        Lianhua~Ji,~
        Chunhua~Chen,~
        Xin~Jing,~
        Ling~Ding,~ \\
        Bingqing~Qu,~
        Pengyang~Wang,~
        Dingqi~Yang*
\IEEEcompsocitemizethanks{\IEEEcompsocthanksitem Bangchao Deng, Lianhua Ji, Chunhua Chen, Xin Jing, Ling Ding, Pengyang Wang, and Dingqi Yang are with the State Key Laboratory of Internet of Things for Smart City and Department of Computer and Information Science, University of Macau, Macao SAR, China, E-mail: \{yc37980, mc45079, yc57438, yc27431, mc36510, pengyangwang, dingqiyang\}@um.edu.mo. Bingqing Qu is with Beijing Normal University-Hong Kong Baptist University United International College, China, E-mail: bingqingqu@uic.edu.cn. \protect\\
\IEEEcompsocthanksitem *Corresponding author: Dingqi Yang (email: dingqiyang@um.edu.mo)}
\thanks{Manuscript received April 19, 2005; revised August 26, 2015.}}

%
%

\markboth{Journal of \LaTeX\ Class Files,~Vol.~14, No.~8, August~2015}%
{Shell \MakeLowercase{\textit{et al.}}: Bare Demo of IEEEtran.cls for Computer Society Journals}
%



\IEEEtitleabstractindextext{%
\begin{abstract}
    Next location prediction is a critical task in human mobility modeling, enabling applications like travel planning and urban mobility management. Existing methods mainly rely on historical spatiotemporal trajectory data to train sequence models that directly forecast future locations. However, they often overlook the importance of the future spatiotemporal contexts, which are highly informative for the future locations. For example, knowing how much time and distance a user will travel could serve as a critical clue for predicting the user's next location. Against this background, we propose \textbf{STRelay}, a universal \textbf{\underline{S}}patio\textbf{\underline{T}}emporal \textbf{\underline{Relay}}ing framework explicitly modeling the future spatiotemporal context given a human trajectory, to boost the performance of different location prediction models. Specifically, STRelay models future spatiotemporal contexts in a relaying manner, which is subsequently integrated with the encoded historical representation from a base location prediction model, enabling multi-task learning by simultaneously predicting the next time interval, next moving distance interval, and finally the next location. We evaluate STRelay integrated with five state-of-the-art location prediction base models on four real-world trajectory datasets. Results demonstrate that STRelay consistently improves prediction performance across all cases by 2.49\%-11.30\%. Additionally, we find that the future spatiotemporal contexts are particularly helpful for entertainment-related locations and also for user groups who prefer traveling longer distances. The performance gain on such non-daily-routine activities, which often suffer from higher uncertainty, is indeed complementary to the base location prediction models that often excel at modeling regular daily routine patterns.
\end{abstract}

}

\maketitle

\IEEEdisplaynontitleabstractindextext

%
\IEEEpeerreviewmaketitle

\IEEEraisesectionheading{\section{Introduction}\label{sec:introduction}}

\IEEEPARstart With the rapid growth of location-based social media, users are increasingly inclined to share their activities on Location-Based Social Networks (LBSNs), generating rich digital footprints for human mobility studies \cite{yang2014modeling}. A key task in human mobility modeling is predicting a user’s next location based on the user's historical mobility trajectory \cite{noulas2012mining}, which is a fundamental building block supporting various location-based applications, such as travel planning \cite{zhou2021contrastive}, taxi ride-sharing scheduling \cite{ma2013t}, and user behavior patterns analyzing \cite{sun2020go}.

\begin{figure}
    \centering
    \includegraphics[width=1.0\linewidth]{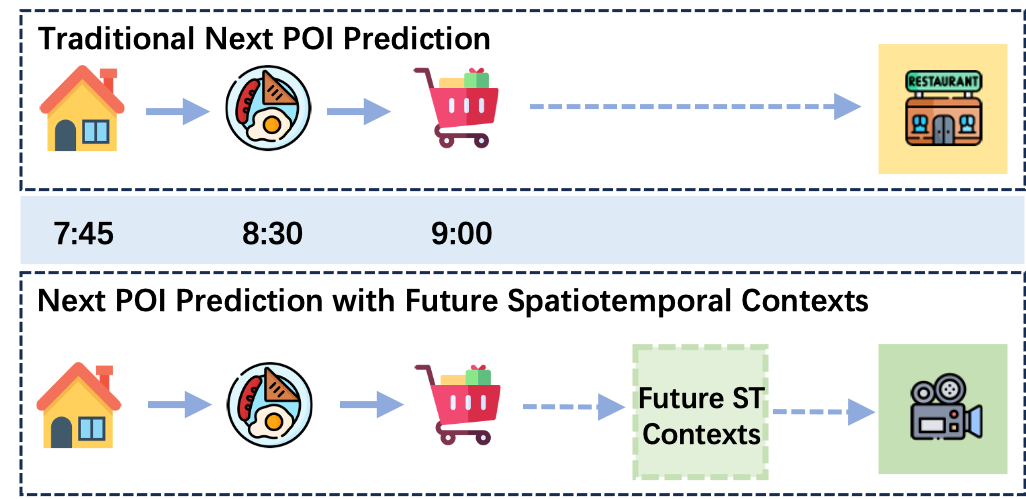}
    \caption{Examples of two prediction paradigms. Paradigm 1 (top): Predict the same POI given historical data of a user. Paradigm 2 (Bottom): Considering future time and moving distance, predict different POIs based on future spatiotemporal contexts.}
    \label{examples}
\end{figure}

To leverage the spatiotemporal characteristics of trajectory data, various methods have been proposed for next Point of Interest (POI) prediction\footnote{Given the context of human trajectories, we do not distinguish the two terms ``location'' and ``POI'' throughout this paper.}. Traditional next POI prediction methods forecast future locations relying solely on historical spatiotemporal information. For example, sequential transitions are typically captured using Markov chain models \cite{rendle2010factorizing, cheng2013you, feng2015personalized}, while long-term dependencies are modeled through RNNs \cite{liu2016predicting, zhao2019go} and attention-based architectures \cite{lian2020geography,liu2024novel,ghosh2024mobilytics,lv2021private}. More recently, some methods using additional context, such as historical spatiotemporal contexts \cite{yang2020location} and social networks \cite{rao2022graph,qin2023diffusion,xu2023revisiting}, have been proposed for further performance improvement. \textit{However, these methods overlook the time information of the predicted next location, and thus always make the same prediction given a historical trajectory}. Some recent approaches have utilized predicted (or even actual) future temporal information to enhance location prediction, effectively capturing time-specific user behaviors. For example, some methods explicitly incorporate future timestamps as query inputs to model time-dependent mobility patterns \cite{feng2024rotan, deng2025replay}. Alternatively, other techniques \cite{yang2022getnext, song2025integrating} jointly optimize timestamp prediction and location forecasting through integrated temporal modeling to improve predictive accuracy. \textit{However, there still lack systematic studies on the key role of future spatiotemporal contexts in location prediction, which, as we show in this paper, are highly informative for the future locations}. For example, knowing how much time and distance a user will travel could serve as a strong indicator for forecasting the user's next location.

Against this background, we propose a novel prediction paradigm by first learning to predict the future spatiotemporal contexts, which are then fed as additional inputs to a base location prediction model to boost the location prediction performance. This paradigm has the following two-fold advantages. On the one hand, human mobility patterns exhibit inherent heterogeneity. For instance, entertainment activities and daily routines (e.g., home-workplace commuting) demonstrate distinct spatiotemporal patterns: the former often exhibits more variable spatiotemporal distributions, while the latter tends to be less variable. By modeling future spatiotemporal contexts, we can effectively accommodate such heterogeneity, in particular, benefiting the more variable entertainment activities, where traditional sequence models often fail (as evidenced by our experiments later). On the other hand, the spatiotemporal preferences of users also exhibit different characteristics. For example, two users might have the same lunch schedule, but their travel preferences differ: a convenience-oriented user may take 10 minutes to go to nearby places like an office cafeteria within 500m; in contrast, an adventurous user may be willing to spend half an hour traveling 5 km to go to a fancy restaurant. Subsequently, modeling the future spatiotemporal contexts of users can effectively benefit such a scenario. Our empirical analysis also shows that incorporating the future spatiotemporal contexts can significantly reduce mobility entropy \cite{zhang2022beyond,sun2024going}, thereby showing great potential to improve the predictability of the next location (see Figure \ref{fig:entropy} in Section \ref{analysis} for details). 

Motivated by the above observations, we propose \textbf{STRelay}, a universal \textbf{\underline{S}}patio\textbf{\underline{T}}emporal \textbf{\underline{Relay}}ing framework that explicitly learns to model the future spatiotemporal contexts given a human trajectory, to boost the performance of different location prediction models. Specifically, we first define the future spatiotemporal contexts as the elapsed time and distance from the present location to the next location in a trajectory, and then discretize them into temporal and spatial intervals under a predefined granularity (1 hour for time or 1 km for distance, respectively). Our objective is to predict the future temporal and spatial intervals based on user trajectories. To capture the inherent relationship between the temporal and spatial contexts, STRelay first learns user- and timestamp-specific representations to query all temporal intervals to obtain a temporal context representation, and subsequently queries all spatial intervals conditioned on the temporal context representation to output a spatial context representation, in a relaying manner. Afterwards, a base location prediction model is used to encode the past trajectory information, which is concatenated with our temporal and spatial context representations to predict the next location. Additionally, to ensure the quality of the learnt future spatiotemporal contexts, we provide supervision signals to the temporal and spatial intervals in the training process under a multi-task learning scheme, where we optimize not only for the prediction of the next location, but also the prediction of the future temporal and spatial intervals.

Our contributions can be summarized as follows:
\begin{itemize}
    \item We reveal the importance of future spatiotemporal contexts for next location prediction tasks, serving as a strong indicator for improving the predictability of the next location.
    \item We propose STRelay, a universal framework designed to explicitly model the future spatiotemporal contexts in a relaying manner; it can be flexibly integrated with different location prediction models under encoder-decoder architectures to boost the prediction performance.
    \item We evaluate STRelay integrated with five state-of-the-art location prediction base models on four real-world trajectory datasets. Results demonstrate that STRelay consistently improves prediction performance across all cases by 2.49\%-11.30\%. Additionally, we find that the future spatiotemporal contexts are particularly helpful for entertainment-related locations and also for user groups who prefer traveling longer distances, which is indeed complementary to the base location prediction models that often excel at modeling daily routine patterns. 
\end{itemize}

\section{Related Work}
Predicting an individual's next location is a fundamental task in human mobility modeling, which aims to predict a user's future location based on their historical mobility trajectories. 
Earlier approaches primarily exploit diverse mobility patterns, such as personal activity preferences \cite{ye2013s,yang2015modeling}, historical visit frequency \cite{noulas2012mining,gao2012exploring}, social ties \cite{cho2011friendship,sadilek2012finding}, and POI or trajectory embeddings learned through graph embedding techniques \cite{xie2016learning,feng2017poi2vec,qian2019spatiotemporal,yang2019revisiting}. Nevertheless, these methods generally fall short in capturing the sequential dynamics of human movement, a factor repeatedly shown to be critical for achieving high-accuracy next-location prediction \cite{liu2016predicting}. 

To more effectively capture the sequential patterns that traditional methods fail to model, sequence-based approaches shift the focus to individual trajectories, explicitly modeling transition patterns across successive POIs in check-in sequences. Early efforts relied heavily on Markov Chain (MC) models to predict the next POI by estimating location-transition probabilities. For example, FPMC \cite{rendle2010factorizing} integrates Markov chains with matrix factorization to derive personalized transition matrices; this paradigm was later augmented with spatial constraints \cite{cheng2013you,feng2015personalized}. Despite their effectiveness on short-range dependencies, these methods struggle to capture long-term sequential patterns in human mobility trajectories. To address long-range dependencies, Recurrent Neural Networks (RNNs, LSTMs, GRUs) and Transformer architectures \cite{vaswani2017attention} have become dominant in next-location prediction. Given that real-world trajectories are typically sparse and irregular, recent models increasingly incorporate spatiotemporal contexts into recurrent units. A widely adopted strategy is to feed temporal and spatial intervals between consecutive check-ins as auxiliary inputs. For instance, Distance2Pre \cite{cui2019distance2pre} conditions predictions on pairwise geographic distances; STRNN \cite{liu2016predicting} constructs time- and space-specific transition matrices to enrich RNN hidden states; HST-LSTM \cite{kong2018hst} extends LSTM gates with spatiotemporal distance; STGN \cite{zhao2019go} capture sequential patterns of users by introducing dedicated spatial and temporal gates into LSTM cells; NeuNext \cite{zhao2020go} jointly optimizes location-context prediction and next-location prediction; STAN \cite{luo2021stan} employs spatiotemporal attention to model non-adjacent dependencies; GeoSAN \cite{lian2020geography} combines hierarchical geographic gridding with self-attention mechanisms, and flashback mechanism \cite{yang2020location}  considers the two universal human mobility laws, i.e., spatial regularity and temporal periodicity, for location prediction. Specifically, it enhances prediction accuracy by leveraging historical hidden states with similar spatiotemporal contexts, inspiring subsequent works \cite{cao2021attention,li2021location,wu2022have,liu2022real,rao2022graph}.

Beyond recurrent architectures, graph-based methods adopt a global perspective by jointly modeling all users’ trajectories. This strategy captures long-range sequential regularities that are difficult to discover from individual sequences alone and enhances POI representations by incorporating influences from locations outside a user’s current trajectory. For example, STP-UDGAT \cite{lim2020stp} applies Graph Attention Networks (GAT) on manually constructed spatial-proximity and frequency-transition graphs to capture both local and global POI correlations;  SGRec \cite{li2021discovering} builds dynamic transition graphs by connecting each POI to its successively visited neighbors and employs GAT for representation learning; Graph-Flashback \cite{rao2022graph} refines POI embeddings via a spatiotemporal knowledge graph; AGRAN \cite{wang2023adaptive} jointly optimizes an adaptive graph structure alongside the sequential predictor in an end-to-end manner; SNPM \cite{yin2023next} introduces a dynamic neighbor graph and models timestamp-level dependencies to better capture transition patterns; AGCL \cite{rao2024next} constructs multiple adaptive graphs (each reflecting a distinct type of POI dependency) and fuses their outputs into a comprehensive multi-faceted POI representation; MobGT \cite{xu2023revisiting} simultaneously encodes user mobility preferences and intrinsic location attributes to extract rich spatial-temporal features; GETNext \cite{yang2022getnext} builds global transition graphs from collective mobility patterns and leverages GNNs for POI representation learning; Diff-POI \cite{qin2023diffusion} exploits transition and distance graphs to derive spatiotemporal embeddings and introduces a diffusion-based sampling strategy to model spatial preferences; and LoTNext \cite{xu2024taming} employs a Long-Tailed Graph Adjustment module to mitigate the impact of long-tailed nodes within the User-POI Interaction Graph. MCLP \cite{sun2024going} employs Latent Dirichlet Allocation (LDA) to discover latent user preferences and leverages a Transformer architecture to model sequential transition patterns.

More recently, researchers have begun leveraging the powerful reasoning and generation capabilities of Large Language Models (LLMs) for human mobility prediction. For example, LLM-Move \cite{feng2024move} and AgentMove \cite{feng2025agentmove} reformulate next-location prediction as a text generation task by converting check-in sequences into natural-language prompts and directly querying frozen LLMs such as ChatGPT; LLMob \cite{wang2024large} employs an agent-based framework that simulates user motivations and activity patterns to generate synthetic mobility trajectories for different personas; MobilityLLM \cite{gong2024mobility} replaces the conventional Transformer backbone with an LLM and fine-tunes it on multiple mobility analysis tasks; LLM4POI \cite{li2024large} constructs an instruction-tuning dataset from context-enriched mobility records (including POI categories and similar historical sequences) to fine-tune LLMs; GenUP \cite{wongso2024genup} further injects explicit user profiles into system prompts to model individual personality traits; QT-Mob \cite{chen2025enhancing} overcomes the limitations of ID-based representations and shallow LLM adaptation, achieving state-of-the-art performance across mobility benchmarks; SILO \cite{sun2025silo} builds a hybrid semantic space that seamlessly integrates traditional ID embeddings, context-aware semantics, and auxiliary information, thereby enabling joint modeling of sequential patterns and rich contextual knowledge. Note that LLM-based approaches are beyond the scope of this paper, as our focus remains on dedicated neural architectures designed specifically for future spatiotemporal contexts modeling.

Furthermore, several methods incorporate predicted or even actual future temporal information to guide location predictions, capturing time-specific user behaviors. For example, they utilize future timestamps as query timestamps \cite{gao2012exploring,yang2019revisiting,li2021location,feng2024rotan,deng2025replay} or a query time interval \cite{zhao2019go,zhao2020go,feng2018deepmove} to predict a user's location after a specified period. Others employ a fixed query time threshold to forecast locations within an upcoming time window \cite{feng2015personalized,feng2018deepmove}. An alternative approach integrates timestamp prediction with location forecasting to improve predictive accuracy \cite{yang2022getnext,sun2024going,song2025integrating}. However, there still lack systematic studies on the key role of future spatiotemporal contexts in location prediction, which, as we show in this paper, are highly informative for the future locations.

Against this background, we propose STRelay, a universal framework that learns to model the future spatiotemporal contexts given a human trajectory, to boost the performance of different location prediction models.

\section{Preliminaries}

In this section, we introduce several preliminaries, including the problem definition and empirical data analyses supporting our motivation.

\subsection{Problem Definition}
\noindent \textbf{Human Trajectory}. A trajectory is a time-ordered sequence $X = \{x_1, x_2, \dots, x_n\}$, where each event (check-in) $x_i = (u, t_i, l_i)$ represents a presence event of a user $u$, with $t_i$ denoting the event time and $l_i$ indicating the location (POI) with its geographic coordinates.

\noindent \textbf{Next Location Prediction}. Given a user's historical trajectory $X = \{x_1, x_2, \dots, x_n\}$, the task aims to predict the most likely POI $l_{n+1}$ that the user will visit in the next check-in event $x_{n+1}$.

\subsection{Empirical Analysis on Future Spatiotemporal Contexts}
\label{analysis}
We conduct an empirical analysis from the perspective of mobility entropy to reveal the importance of the future spatiotemporal context for next location prediction. Specifically, we first focus solely on the trajectory sequence and compute the mobility entropy \cite{zhang2022beyond} for each user as follows:
\begin{equation}
E(S^{u}) = -\sum_{i=1}^{N^{u}} p(l_i) \log_2 p(l_i)
\label{eq:entropy}
\end{equation}
where $S^{u}$ represents the trajectory sequence of user $u$, $N^{u}$ denotes the number of unique locations visited by $u$ in $S^{u}$, and $p(l_i)$ denotes the frequency of location $l_i$ occurring in $S^{u}$.

\begin{figure}[]
    \subfigure[Istanbul]{
        \includegraphics[width=0.47\linewidth]{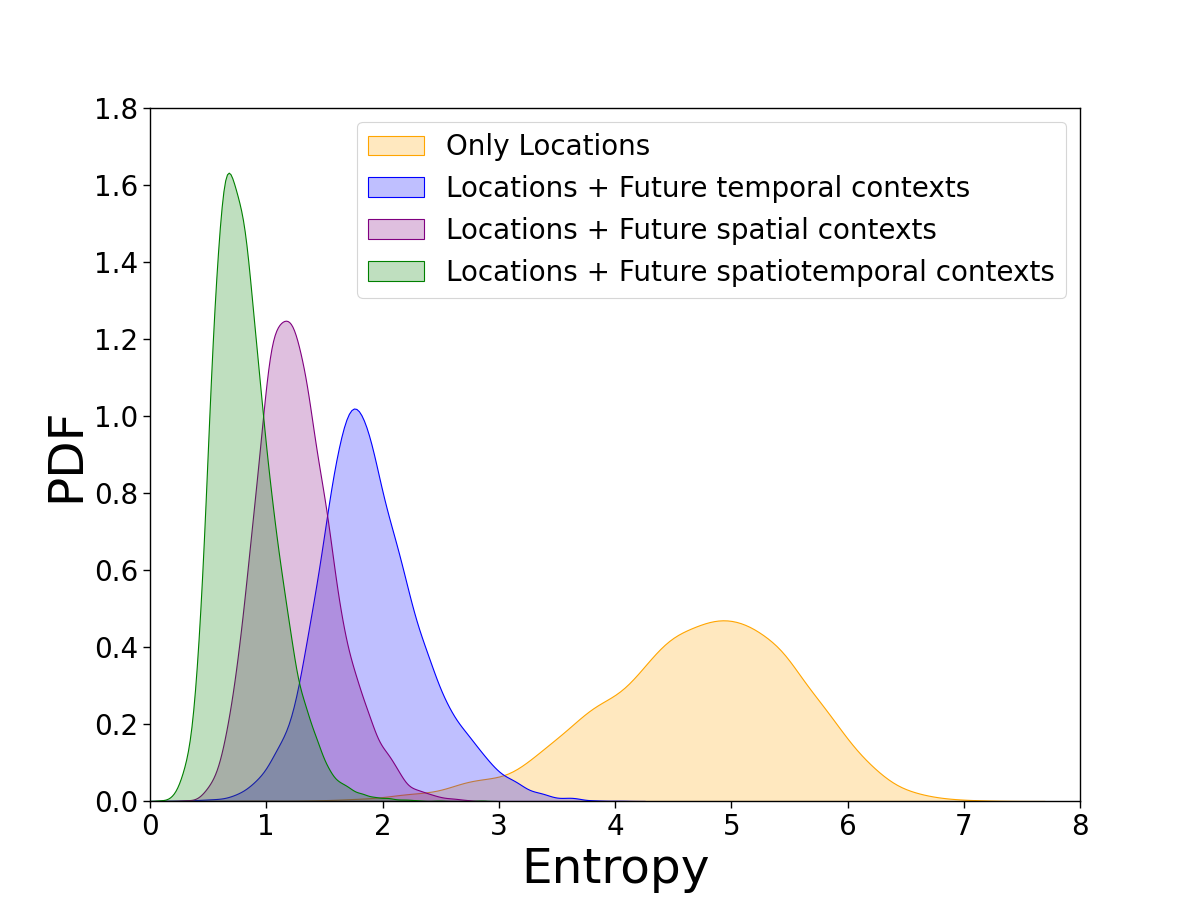} 
        \label{fig:ist_entropy}}
    \subfigure[Tokyo]{
        \includegraphics[width=0.47\linewidth]{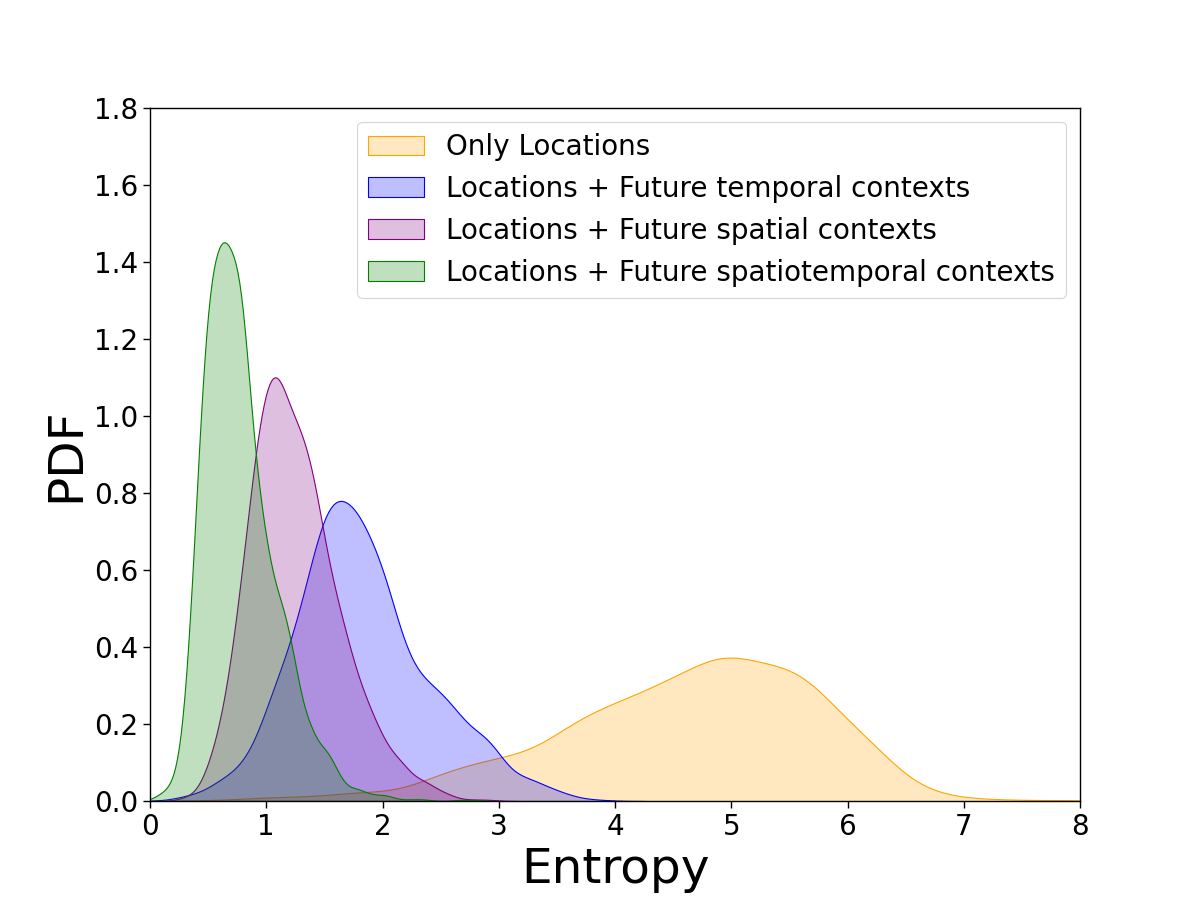} 
        \label{fig:tky_entropy}}
    \subfigure[Singapore]{
        \includegraphics[width=0.47\linewidth]{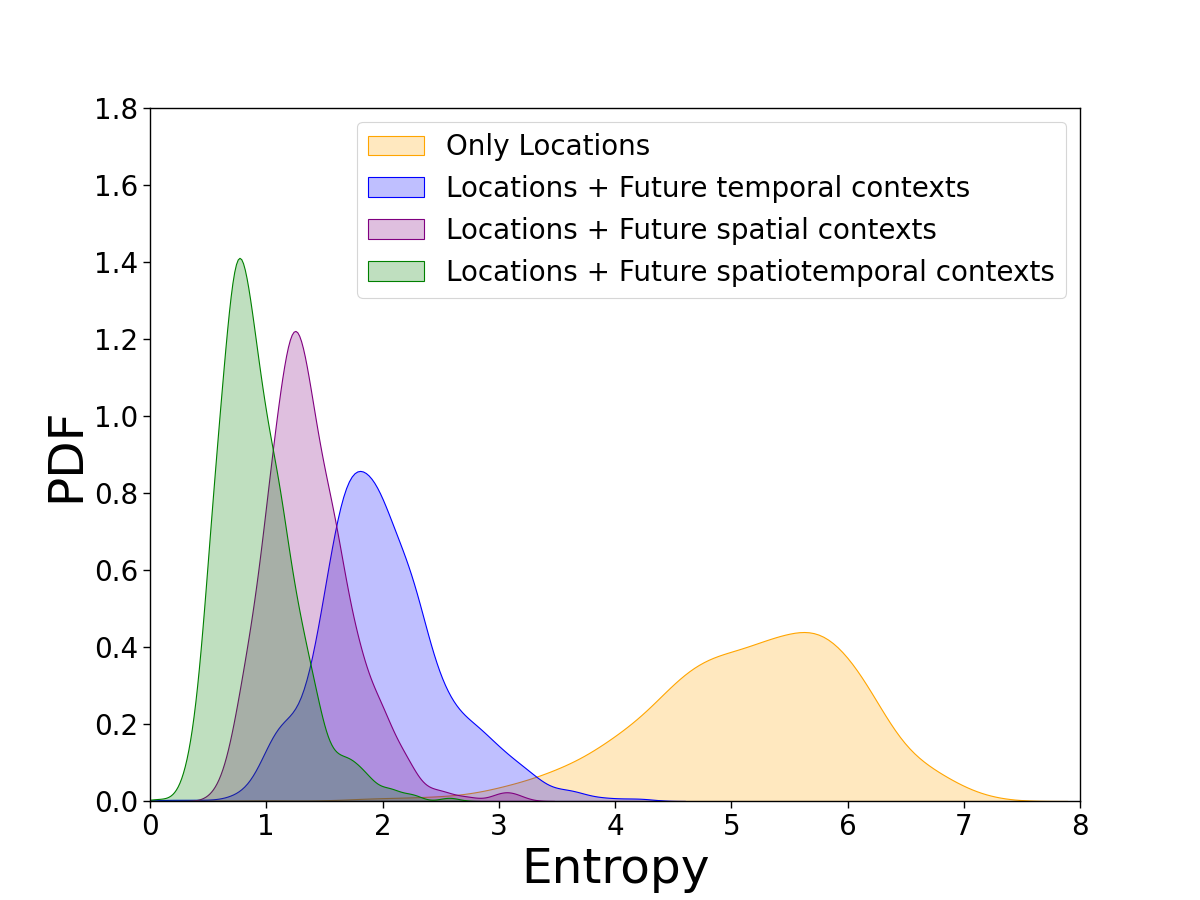} 
        \label{fig:sin_entropy}}
    \subfigure[Moscow]{
        \includegraphics[width=0.47\linewidth]{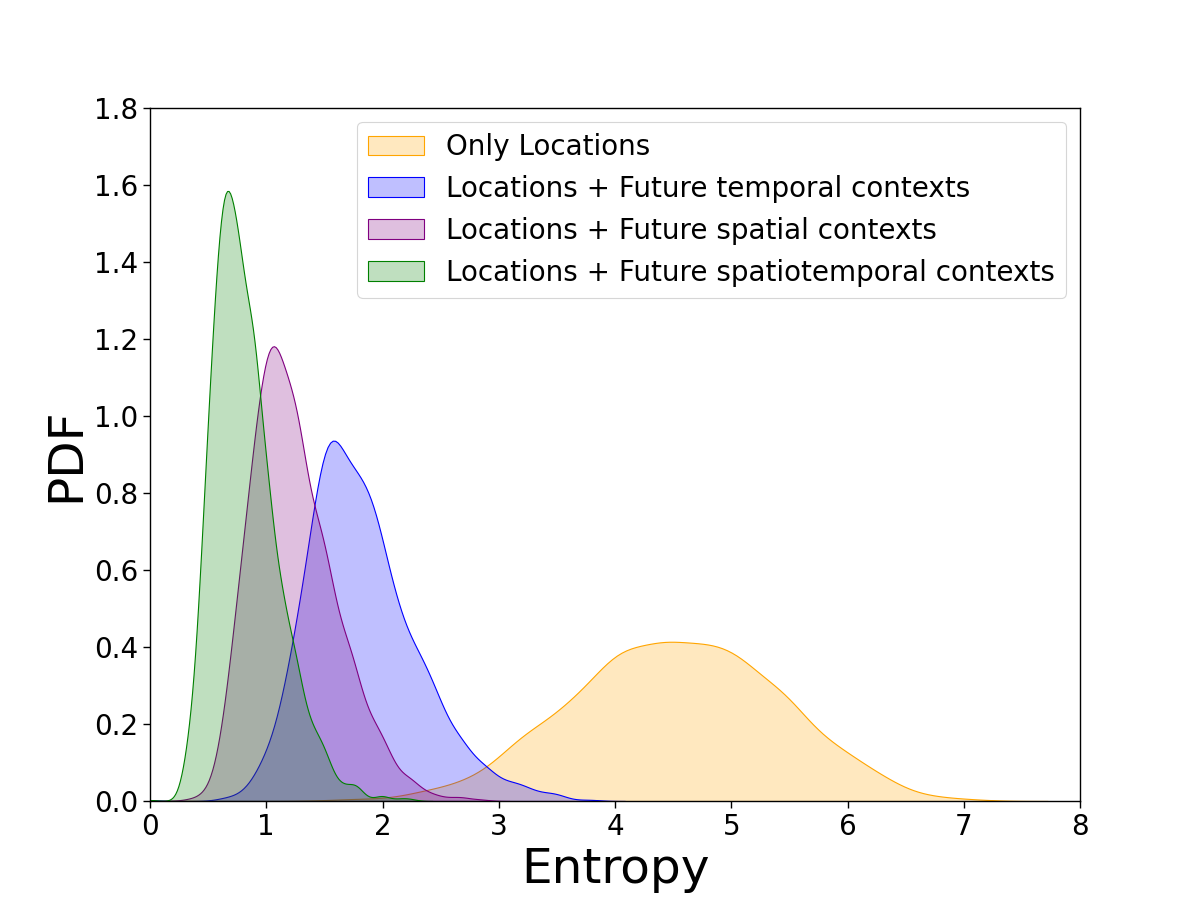} 
        \label{fig:mos_entropy}}
    \caption{Influence of spatiotemporal contexts on mobility entropy}
    \label{fig:entropy}
\end{figure}

Furthermore, we introduce future temporal context \cite{sun2024going} into the mobility entropy through categorizing locations visited by users into their respective future $t$, and then compute the mean mobility entropy after accounting for future temporal context as follows:
\begin{equation}
E^t(S^{u}) = -\frac{1}{N^{t}} \sum_{{t}=1}^{N^{t}} \sum_{i=1}^{N_t^{u}} p(l_i,{t}) \log_2 p(l_i,{t})
\label{eq:entropy_t}
\end{equation}
where $N^{t}$ represents the number of $t$, $N_{t}^{u}$ is the count of unique locations visited by user $u$ with $t$, and $p(l_i,t)$ denotes the frequency of location $l_i$ appearing with $t$.

Similarly, we introduce future spatial context into the mobility entropy through categorizing locations visited by users into their respective future moving distance $d$, and then compute the mean mobility entropy after accounting for future spatial context as follows:
\begin{equation}
E^s(S^{u}) = -\frac{1}{N^{d}} \sum_{{d}=1}^{N^{d}} \sum_{i=1}^{N_d^{u}} p(l_i,{d}) \log_2 p(l_i,{d})
\label{eq:entropy_s}
\end{equation}
where $N^{d}$ represents the number of $d$, $N_{d}^{u}$ is the count of unique locations visited by user $u$ with $d$, and $p(l_i,d)$ denotes the frequency of location $l_i$ appearing with $d$.

In the end, we further consider both spatial and temporal contexts to get mobility entropy with the future spatiotemporal contexts:
\begin{equation}
E^{st}(S^{u}) = -\frac{1}{N^t} \frac{1}{N^{d}} \sum_{t=1}^{N^t} \sum_{{d}=1}^{N^{d}} \sum_{i=1}^{N_{td}^{u}} p(l_i,{t},d) \log_2 p(l_i,{t},d)
\label{eq:entropy_st}
\end{equation}
where $N^d$ represents the number of distance intervals, $N_{td}^{u}$ is the count of unique locations visited by user $u$ with $t$ in $d$, and $p(l_i,t,d)$ denotes the frequency of location $l_i$ appearing with $t$ in $d$.

The distributions of the mobility entropy on four datasets are shown in Figure \ref{fig:entropy}. We observe that the introduction of temporal context or spatial context can effectively reduce mobility entropy in both datasets. This reduction suggests that involving future spatiotemporal contexts for next location prediction has great potential to reduce the uncertainty of the next location, ultimately resulting in higher predictability.

\section{STRelay}

\begin{figure}
    \centering
    \includegraphics[width=0.95\linewidth]{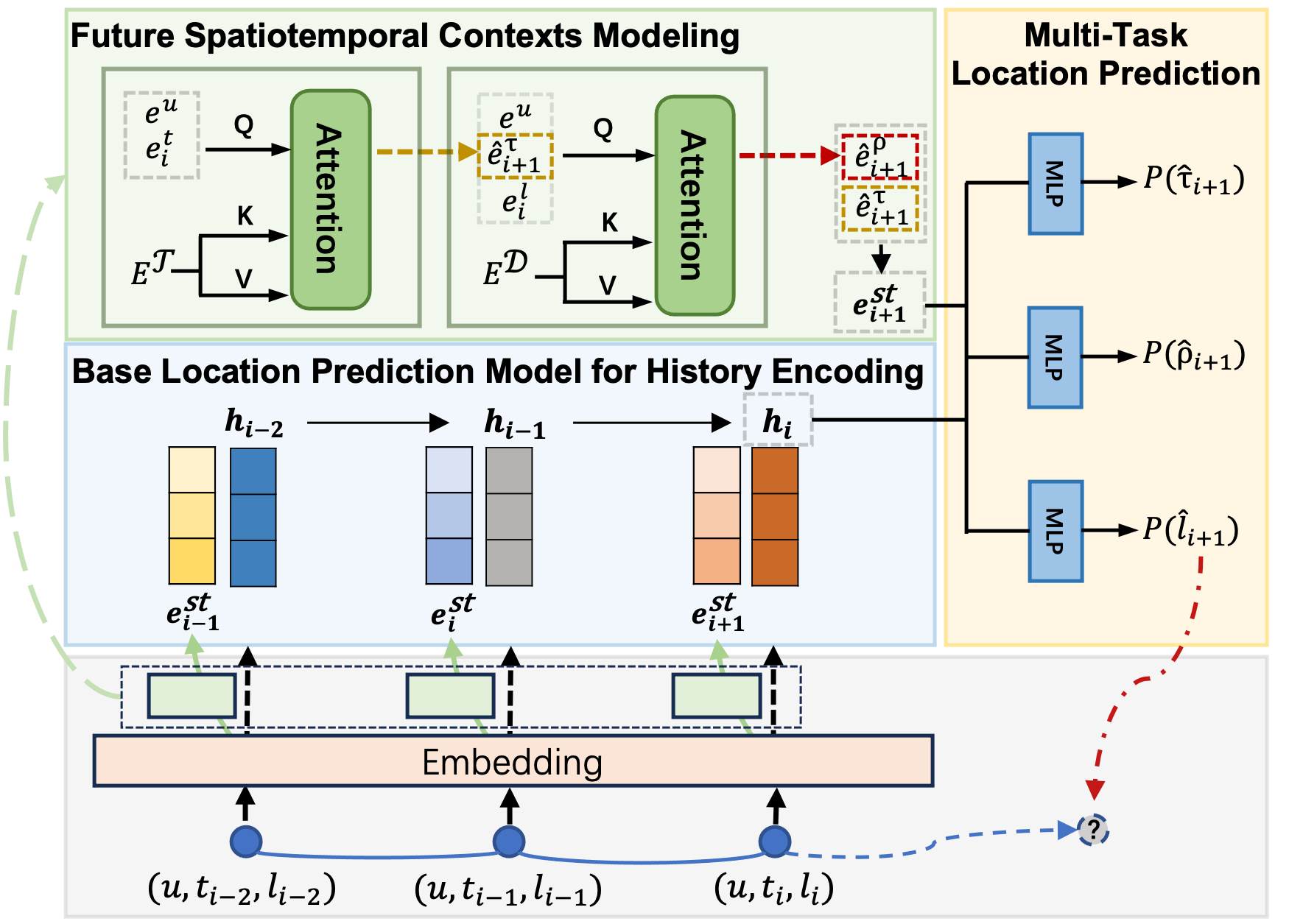}
    \caption{Our proposed STRelay consists of 1) a future spatiotemporal contexts modeling module, 2) a base location prediction model for history encoding, and 3) a multi-task prediction module.}
    \label{model_arxiv}
\end{figure}

The framework of our STRelay is shown in Figure \ref{model_arxiv}. First, to model future spatiotemporal contexts to boost the performance, we discretize spatiotemporal contexts into temporal and spatial intervals under a predefined granularity. Then, we capture the inherent relationship between temporal and spatial contexts by first learning to model temporal context representations, and subsequently conditioning on the temporal context representation to output a spatial context representation. Afterward, we encode historical latent representation from a base location prediction model, along with our spatial and temporal context representations, and feed them into a multi-task prediction layer, simultaneously predicting the next time interval, next moving distance interval, and finally the next location. In the following, we elaborate on our model design, followed by the training process.

\subsection{Future Spatiotemporal Context Modeling}

In this paper, we define the future spatiotemporal contexts as the elapsed time and distance from a user's current location to their next location in a trajectory. By effectively modeling how much time the user will spend traveling and how far the user will go, such spatiotemporal contexts serve as critical predictive features for next location forecasting, providing deeper insights into users' forthcoming mobility patterns, thereby enhancing location prediction performance.

\subsubsection{Discretization of Future Spatiotemporal Contexts.} The future spatiotemporal contexts are inherently continuous and dynamic; we first discretize them into temporal and spatial intervals under a predefined granularity. 

For temporal intervals, we define a minimum interval $\Delta t$ (e.g., 1 hour) and a maximum interval $M\Delta t$ (e.g., 24 hours), constructing a discrete candidate set: 
\begin{align}
    \mathcal{T} = \{\Delta t, 2\Delta t, \ldots, M\Delta t\}
\end{align}
Noted that any interval exceeding $M\Delta t$ is capped at $M\Delta t$. Each temporal interval between consecutive check-ins is assigned to its corresponding bin and represented as an $M$-dimensional one-hot vector $\tau \in \mathbb{R}^{M}$.

Similarly, for spatial intervals, we define a minimum interval $\Delta d$ (e.g., 1 km) and a maximum interval $N\Delta d$ (e.g., 30 km), forming a discrete candidate set: 
\begin{align}
    \mathcal{D} = \{\Delta d, 2\Delta d, \ldots, N\Delta d\}
\end{align}
where any distances exceeding $N\Delta d$ are capped at $N\Delta d$, and each future spatial distance between consecutive check-ins is assigned to its corresponding bin and represented as an $N$-dimensional one-hot vector $\rho \in \mathbb{R}^{N}$. Afterwards, for all candidate future temporal and spatial intervals, we embed them into a latent space as learnable embeddings $E^{\mathcal{T}} \in \mathbb{R}^{M \times d}$ and $E^{\mathcal{D}} \in \mathbb{R}^{N \times d}$, respectively, where $d$ is the dimension of the embedding space. 

These discretization and embedding approaches enhance training stability by reducing sensitivity to minor variations (e.g., treating 6 hours and 6.17 hours, or 8 km and 8.3 km as the same intervals since they share similar properties), enabling robust modeling of temporal and spatial context representations for next POI prediction. We also conduct experiments to investigate the impact of different temporal and spatial granularities in our experiments later.

\subsubsection{Representation of Future Spatiotemporal Contexts.} 
Human mobility exhibits strong spatiotemporal coupling, where daily routines (e.g., home-work commutes) typically involve both short time intervals and moving distances, while entertainment activities often exhibit longer travel times and distances. To capture such inherent spatiotemporal correlation, we adopt a relaying manner that generates future spatiotemporal context representations by mimicking an individual's sequential decisions about how much time and how far to move next \cite{yuan2022activity,deng2025revisiting}. Specifically, we first learn a temporal context representation and subsequently condition on the learnt temporal context representation to output a spatial context representation.

Intuitively, the temporal context for the next check-in depends on the user and the current time. For example, a user at a workplace on a weekday morning may check in at a nearby café within 1 hour, while the same user at a park on a weekend may check in at a distant restaurant after 5 hours. To capture this dependency, when predicting next location $l_{i+1}$, we first embed the user $u$ into the low-dimensional embedding space, denoted as $e^u \in \mathbb{R}^d$. Next, for the current timestamp $t_i$, we convert it to an hour-in-week representation \cite{deng2025replay} and embed it into a latent space as a learnable embedding $e^t_i \in \mathbb{R}^d$, which can naturally capture temporal regularities between weekdays and weekends. Then, we concatenate them to get the temporal query embedding $e^{\tau}_i$ for predicting future temporal context:
\begin{equation}
    e^{\tau}_i = [e^u; e^t_i]
\end{equation}
Then, we use temporal query embedding to query all of the candidate temporal interval embeddings $E^{\mathcal{T}}$ via an attention mechanism \cite{vaswani2017attention}, producing a future temporal context representation $\hat{e}^{\tau}_{i+1}$:
\begin{flalign}
    & Q = e^{\tau}_i W_Q; K  = E^{\mathcal{T}} W_K ; V = E^{\mathcal{T}} W_V \label{qkv} \\ 
    & \hat{e}^{\tau}_{i+1} = \text{softmax}\left(\frac{QK^T}{\sqrt{d_k}}\right)V  \label{attn}
\end{flalign}
where $W_Q$, $W_K$, and $W_V$ are the learnable linear projection matrices to map the input into a corresponding subspace.

After obtaining the future temporal representation, we now turn to modeling the spatial dimension. Generally, how far a user will travel next is influenced by her personal preference, the anticipated timing (i.e., the future temporal representation), and the current location information. To encode these spatiotemporal dependencies and model the future spatial context, we construct a spatial query embedding $e^{\rho}_i$ by concatenating three components: the user embedding $e^u$, the future temporal representation $\hat{e}^{\tau}_{i+1} $, and the latent embedding of the current location $e^l_i \in \mathbb{R}^d$,  formulated as:
\begin{equation}
    e^{\rho}_i = [e^u; \hat{e}^{\tau}_{i+1}; e^l_i]
\end{equation} 

Afterwards, we get future spatial context representation in a similar way to the temporal aspect as shown in Eq. \ref{qkv} and Eq. \ref{attn}. Specifically, we use spatial query embedding $e^{\rho}_i$ as query and $E^{\mathcal{D}}$ as key and value to produce a weighted future spatial representation $\hat{e}^{\rho}_{i+1}$ by an attention mechanism. 

To this end, we concatenate the future temporal context representation $\hat{e}^{\tau}_{i+1}$ and future spatial context representation $\hat{e}^{\rho}_{i+1}$ to obtain the future spatiotemporal context representations $e^{st}_{i+1}$.

\subsection{History Encoding and Location Prediction}
\subsubsection{Base Location Prediction Model for History Encoding} 
Our STRelay framework is designed to flexibly integrate with different location prediction models under encoder-decoder architectures. While $e^{st}_{i+1}$ captures the future temporal and spatial contexts, to predict the next POI $l_{i+1}$, we also need to encode historical trajectory information. Given a base location prediction model $g$, we input the historical trajectory $X$ to obtain $h_i \in \mathbb{R}^{d}$, which encodes the historical sequential spatiotemporal information until $t_i$:
\begin{equation}
    h_i = g (X)
\end{equation}
Subsequently, we concatenate the hidden state of the trajectory history $h_i$ and the predicted future spatiotemporal context representation $e^{st}_{i+1}$ as the overall context embedding $e^c$, which integrates both historical and future spatiotemporal contexts for prediction:
\begin{equation}
    e^c = [h_i; e^{st}_{i+1} ]
\end{equation}

\subsubsection{Multi-Task Location Prediction} 
After the overall context embedding is generated, it is then fed into a multi-layer perceptron (MLP) followed by a softmax function to predict the next location:
\begin{align} 
& P(\hat{l}_{i+1}) = \text{softmax}(MLP_{\theta} (e^c))
\end{align}
where $\hat{l}_{i+1}$ is the predicted probability distributions for next location.

Meanwhile, to ensure the quality of the learnt future spatiotemporal contexts, we provide supervision signals to the temporal and spatial intervals in the training process under a multi-task learning scheme alongside the POI prediction, using the context embedding $e^c$. We define two prediction heads to estimate the probability distributions of future temporal and spatial intervals:
\begin{align} 
 & P(\hat{\tau}_{i+1}) = \text{softmax}(MLP_{\gamma} (e^c)), \\ 
 & P(\hat{\rho}_{i+1}) = \text{softmax}(MLP_{\phi} (e^c))
\end{align} 
where $\hat{\tau}_{i+1}$ and $\hat{\rho}_{i+1}$ are the predicted probability distributions for the future temporal and spatial intervals, respectively. 

\subsection{Model Training}
The training objective of STRelay can be decomposed into a base location prediction loss and two future spatiotemporal context losses. First, the base historical modeling loss aims to predict the next location:
\begin{flalign}
    \mathcal{L}_{\text{POI}} = - \sum^{U}_{j=1} \sum^{N_j-1}_{i=1} l_{i+1,j} log(P(\hat{l}_{i+1,j}))
\end{flalign}
where $U$ is the number of users; $N_j$ is the sequence length of $j$-th user; $l_{i+1,j}$ is the one-hot vector of ground truth POI for ($i$+$1$)-th check-in of $j$-th user. 

Since both temporal and spatial contexts are discretized, their losses also take the form of a multi-class classification loss. We get the loss of predicting future temporal and spatial contexts as follows:
\begin{flalign}
    \mathcal{L}_{\tau} = - \sum^{U}_{j=1} \sum^{N_j-1}_{i=1} \tau_{i+1,j} log(P(\hat{\tau}_{i+1,j})) \\
    \mathcal{L}_{\rho} = - \sum^{U}_{j=1} \sum^{N_j-1}_{i=1} \rho_{i+1,j} log(P(\hat{\rho}_{i+1,j}))
\end{flalign}

The overall loss function of our STRelay combines the three losses as follows:
\begin{flalign}
    \mathcal{L} = \mathcal{L}_{\text{POI}}+ \mathcal{L}_{\tau} +\mathcal{L}_{\rho}
\end{flalign}

\section{Experiments}
\begin{table}[]
    \centering
    \caption{Datasets Statistics}
    \label{data_statistic}
    \fontsize{8.5pt}{10pt}\selectfont
    \begin{tabular}{c|ccc}
        \hline
        Dataset&  \#Users & \#POIs &  \#Check-ins  \\ \hline
        Istanbul&8,750 &13,194 & 1,530,050\\ 
        Tokyo&2,484 &8,196 & 735,211\\ 
        Singapore&1,164 &10,859 &281,039 \\ 
        Moscow&2,665 &12,527 & 516,355\\ \hline
    \end{tabular}
\end{table}

\begin{table*}[!htp]\centering
\caption{Performance Comparison of Baselines and STRelay-Enhanced Methods on Istanbul and Tokyo Datasets}
\label{tab:all_results_ist_tky}
\fontsize{8.5pt}{10pt}\selectfont
\setlength{\tabcolsep}{0.5em}
\begin{tabular}{l|ccccc|ccccc} \hline
 & \multicolumn{5}{c|}{Istanbul} & \multicolumn{5}{c}{Tokyo} \\ \hline
Methods &Acc@5 &Acc@10 &NDCG@5 &NDCG@10 &MRR &Acc@5 &Acc@10 &NDCG@5 &NDCG@10 &MRR \\  \hline
BPR &0.2532 &0.3816 &0.1821 &0.2500 &0.1528 &0.2133 &0.3215 &0.1565 &0.2121 &0.1317 \\
Caser &0.3179 &0.3751 &0.2464 &0.2649 &0.2396 &0.4115 &0.5000 &0.2988 &0.3275 &0.2830 \\
STGCN &0.3312 &0.3949 &0.2514 &0.2721 &0.2428 &0.4331 &0.5306 &0.3104 &0.3422 &0.2927 \\
DeepMove &0.2633 &0.3368 &0.1903 &0.2141 &0.1859 &0.4127 &0.5107 &0.3020 &0.3338 &0.2884 \\
STAN &0.3014 &0.3716 &0.2240 &0.2467 &0.2171 &0.3911 &0.4890 &0.2830 &0.3147 &0.2702 \\
GETNext &0.3107 &0.3816 &0.2306 &0.2536 &0.2230 &0.4214 &0.5197 &0.3089 &0.3408 &0.2948 \\ \hline
STGN &0.3292 &0.3936 &0.2495 &0.2704 &0.2410 &0.4415 &0.5342 &0.3228 &0.3529 &0.3056 \\
Flashback &0.3475 &0.4139 &0.2632 &0.2847 &0.2557 &0.4533 &0.5459 &0.3339 &0.3640 &0.3119 \\
Graph-Flashback &0.3521 &0.4173 &\underline{0.2711} &\underline{0.2925} &0.2578 &\underline{0.4589} &\underline{0.5548} &0.3346 &0.3647 &0.3154 \\
SNPM &0.3458 &0.4125 &0.2633 &0.2849 &0.2527 &0.4538 &0.5470 &0.3305 &0.3607 &0.3109 \\
LoTNext &\underline{0.3530} &\underline{0.4199} &0.2684 &0.2901 &\underline{0.2587} &0.4581 &0.5534 &\underline{0.3350} &\underline{0.3660} &\underline{0.3165} \\ \hline
STRealy+STGN &0.3604 &0.4316 &0.2690 &0.2921 &0.2576 &0.4625 &0.5569 &0.3412 &0.3709 &0.3232 \\
STRealy+Flashback &0.3779 &0.4473 &0.2847 &0.3072 &0.2721 &0.4803 &0.5743 &0.3552 &0.3859 &0.3377 \\
STRealy+SNPM &0.3781 &0.4484 &0.2846 &0.3070 &0.2723 &0.4801 &0.5786 &0.3521 &0.3837 &0.3329 \\
STRealy+LoTNext &0.3744 &0.4455 &0.2821 &0.3051 &0.2704 &0.4856 &0.5828 &\textbf{0.3612} &\textbf{0.3929} &\textbf{0.3423} \\
STRealy+Graph-Flashback &\textbf{0.3807} &\textbf{0.4525} &\textbf{0.2873} &\textbf{0.3101} &\textbf{0.2743} &\textbf{0.4863} &\textbf{0.5835} &0.3555 &0.3862 &0.3381 \\\hline
 
\end{tabular}
\end{table*}

\begin{table*}[!htp]\centering
\caption{Performance Comparison of Baselines and STRelay-Enhanced Methods on Singapore and Moscow Datasets}
\label{tab:all_results_sin_mos}
\fontsize{8.5pt}{10pt}\selectfont
\setlength{\tabcolsep}{0.5em}
\begin{tabular}{l|ccccc|ccccc} \hline
 & \multicolumn{5}{c|}{Singapore} & \multicolumn{5}{c}{Moscow} \\ \hline
Methods &Acc@5 &Acc@10 &NDCG@5 &NDCG@10 &MRR &Acc@5 &Acc@10 &NDCG@5 &NDCG@10 &MRR \\  \hline
BPR &0.2077 &0.3193 &0.1466 &0.2069 &0.1176 &0.1362 &0.2093 &0.0937 &0.1273 &0.0820 \\
Caser &0.2392 &0.3091 &0.1726 &0.1953 &0.1600 &0.3239 &0.3974 &0.2341 &0.2579 &0.2301 \\
STGCN &0.2917 &0.3691 &0.2102 &0.2353 &0.2138 &0.3552 &0.4237 &0.2587 &0.2810 &0.2437 \\
DeepMove &0.2923 &0.3871 &0.2053 &0.2360 &0.2009 &0.2905 &0.3790 &0.2035 &0.2322 &0.1967 \\
STAN &0.2939 &0.3855 &0.2086 &0.2382 &0.2040 &0.3105 &0.3993 &0.2186 &0.2474 &0.2102 \\
GETNext &0.3171 &0.4008 &0.2293 &0.2564 &0.2217 &0.3528 &0.4302 &0.2580 &0.2831 &0.2456 \\ \hline
STGN &0.2736 &0.3538 &0.1950 &0.2209 &0.1905 &0.3475 &0.4157 &0.2533 &0.2755 &0.2392 \\
Flashback &0.2995 &0.3815 &0.2159 &0.2425 &0.2092 &0.3780 &0.4410 &0.2802 &0.3007 &0.2633 \\
Graph-Flashback &0.3207 &0.4041 &0.2314 &0.2584 &0.2233 &0.3986 &0.4645 &0.2967 &0.3197 &0.2765 \\
SNPM &0.3138 &0.3993 &0.2268 &0.2520 &0.2164 &0.3913 &0.4562 &0.2906 &0.3118 &0.2738 \\
LoTNext &\underline{0.3241} &\underline{0.4093} &\underline{0.2342} &\underline{0.2618} &\underline{0.2260} &\underline{0.4025} &\underline{0.4719} &\underline{0.2990} &\underline{0.3216} &\underline{0.2814} \\ \hline
STRealy+STGN &0.2914 &0.3748 &0.2080 &0.2350 &0.2024 &0.3841 &0.4633 &0.2819 &0.3076 &0.2669 \\
STRealy+Flashback &0.3116 &0.3978 &0.2243 &0.2523 &0.2175 &0.4009 &0.4740 &0.2939 &0.3177 &0.2756 \\
STRealy+SNPM &0.3279 &0.4172 &0.2382 &0.2663 &0.2259 &0.4167 &0.4944 &0.3047 &0.3288 &0.2877 \\
STRealy+LoTNext &0.3312 &0.4188 &0.2402 &0.2686 &0.2323 &0.4192 &0.4949 &0.3092 &0.3338 &0.2905 \\
STRealy+Graph-Flashback &\textbf{0.3348} &\textbf{0.4251} &\textbf{0.2433} &\textbf{0.2728} &\textbf{0.2338} &\textbf{0.4217} &\textbf{0.4970} &\textbf{0.3113} &\textbf{0.3360} &\textbf{0.2915} \\ \hline
 
\end{tabular}
\end{table*}

\subsection{Experiment Setup}
\subsubsection{Dataset}  We evaluate our STRelay using four widely used real-world datasets: Istanbul, Tokyo, Singapore and Moscow. They are collected from April 2012 to February 2013 \cite{yang2019revisiting}. Each check-in consists of the user ID, POI ID, latitude, longitude, and timestamp. We follow the data pre-processing steps in \cite{yang2020location}. The initial 80\% of the data (in chronological order) is used for training, and the remaining data for testing. Table \ref{data_statistic} summarizes the details of the four processed datasets.

\subsubsection{Baselines}
We choose the following models for location prediction as baselines: 
\begin{itemize}
    \item \textbf{BPR} \cite{rendle2012bpr} employs a pairwise ranking loss to learn user preferences; it exploits direct user-item interaction to separate negative items from positive items to alleviate the sparsity of data and improves the performance.
    \item \textbf{Caser} \cite{tang2018personalized} embeds POIs in user interaction history as images, using convolutional filters to capture both local and global sequential dependencies.
    \item \textbf{STGN} \cite{zhao2019go}: add additional gates controlled by the spatiotemporal distances between successive check-ins to LSTM. Note that we add user preference modeling in our experiments for this method.
    \item \textbf{STGCN} \cite{zhao2019go} is a variant of STGN with coupled input and forget gates for improved efficiency. Note that we add user preference modeling in our experiments for this method.
    \item \textbf{DeepMove} \cite{feng2018deepmove} adds an attention mechanism to GRU for location prediction over sparse mobility traces.
    \item \textbf{STAN} \cite{luo2021stan} models relative spatial-temporal information among POIs with a bi-layer attention architecture.
    \item \textbf{GETNext} \cite{yang2022getnext} utilizes a global trajectory flow graph to enhance the prediction of the next POI, alongside proposing a GCN model for generating effective POI embeddings.
    \item \textbf{Flashback} \cite{yang2020location} is an RNN-based framework that explicitly leverages high-order spatiotemporal distance information to identify informative past hidden states.
    \item \textbf{Graph-Flashback} \cite{rao2022graph} constructs a spatial-temporal knowledge graph for POI representation learning.
    \item \textbf{SNPM} \cite{yin2023next} represents POI relationships through knowledge graph embeddings and aggregates contextual information from neighboring nodes.
    \item \textbf{LoTNext} \cite{xu2024taming} employs a Long-Tailed Graph Adjustment module to mitigate the impact of long-tailed nodes within the User-POI Interaction Graph.
\end{itemize}
Among these baselines, we instantiate our STRelay with five techniques: STGN, Flashback, Graph-Flashback, SNPM, and LoTNext to validate its effectiveness.

\subsubsection{Evaluation Metrics}
We adopt two evaluation metrics commonly used in prior research: Mean Reciprocal Rank (MRR), average Accuracy@K (Acc@K), and Normalized Discounted Cumulative Gain@K (NDCG@K) to evaluate the performance. Their definitions are as follows:
\begin{equation}
\text{MRR} = \frac{1}{m} \sum_{i=1}^{m} \frac{1}{rank}
\end{equation}
where $m$ is the number of predictions and the $rank$ represents the rank of the true POI in the predicted ordered list. Acc@K is the rate of true positive samples in the predicted top-$K$ positive samples, which is computed as follows:
\begin{equation}
\text{ACC}@K = \frac{1}{|U|} \sum_{u \in U} \frac{|S_u^K \cap S_u^L|}{|S_u^L|},
\end{equation}
where $S_u^K$ is the set of predicted top-$K$ POIs for user $u$, $S_u^L$ denotes the ground truth of user $u$. Unlike Acc@K, which focuses on top-$K$, NDCG@K aims to measure the overall ranking performance of the model. The calculation formulation of NDCG@K is presented as follows:

\begin{equation}
\text{NDCG}@K = \frac{1}{|U|} \sum_{u \in U} \frac{1}{\log(rank_u + 1)}
\end{equation}
where $rank_u$ denotes the rank of $S_u^L$ in $S_u^K$ for user $u$. In our experiments, we adopt the commonly used $K=\{5,10\}$.

\begin{table*}[!htp]\centering
\caption{Performance Comparison of Different STRelay Variants on Istanbul and Tokyo Datasets}
\vspace{-1em}
\label{tab:ablation_results_ist_tky}
\setlength{\tabcolsep}{0.5em}
\begin{tabular}{l|ccccc|ccccc} \hline
 & \multicolumn{5}{c|}{Istanbul} & \multicolumn{5}{c}{Tokyo} \\ \hline
Methods &Acc@5 &Acc@10 &NDCG@5 &NDCG@10 &MRR &Acc@5 &Acc@10 &NDCG@5 &NDCG@10 &MRR \\  \hline
STRelay+STGN &\textbf{0.3604} &\textbf{0.4316} &\textbf{0.2690} &\textbf{0.2921} &\textbf{0.2576} &\textbf{0.4625} &\textbf{0.5569} &\textbf{0.3412} &\textbf{0.3709} &\textbf{0.3232} \\
\textit{\textit{w/o spatial}} &0.3384 &0.4050 &0.2574 &0.2790 &0.2491 &0.4467 &0.5373 &0.3304 &0.3599 &0.3138 \\
\textit{w/o temporal} &0.3344 &0.4013 &0.2547 &0.2763 &0.2468 &0.4475 &0.5381 &0.3315 &0.3609 &0.3148 \\
\textit{w/o relaying} &0.3418 &0.4095 &0.2599 &0.2819 &0.2515 &0.4510 &0.5437 &0.3332 &0.3633 &0.3161 \\
\textit{w/o spatiotemporal} &0.3292 &0.3936 &0.2495 &0.2704 &0.2410 &0.4415 &0.5342 &0.3228 &0.3529 &0.3056 \\ \hline
STRealy+Flashback &\textbf{0.3779} &\textbf{0.4473} &\textbf{0.2849} &\textbf{0.3073} &\textbf{0.2721} &\textbf{0.4803} &\textbf{0.5743} &\textbf{0.3552} &\textbf{0.3859} &\textbf{0.3377} \\
\textit{w/o spatial} &0.3670 &0.4340 &0.2782 &0.2999 &0.2668 &0.4587 &0.5559 &0.3356 &0.3672 &0.3171 \\
\textit{w/o temporal} &0.3686 &0.4366 &0.2792 &0.3012 &0.2677 &0.4620 &0.5586 &0.3381 &0.3695 &0.3194 \\
\textit{w/o relaying} &0.3726 &0.4404 &0.2810 &0.3029 &0.2688 &0.4728 &0.5651 &0.3506 &0.3807 &0.3316 \\
\textit{w/o spatiotemporal} &0.3475 &0.4139 &0.2632 &0.2847 &0.2557 &0.4533 &0.5459 &0.3339 &0.3640 &0.3119 \\ \hline
STRealy+SNPM &\textbf{0.3781} &\textbf{0.4484} &\textbf{0.2846} &\textbf{0.3070} &\textbf{0.2723} &\textbf{0.4801} &\textbf{0.5786} &\textbf{0.3521} &\textbf{0.3837} &\textbf{0.3329} \\
\textit{w/o spatial} &0.3682 &0.4364 &0.2790 &0.3011 &0.2677 &0.4721 &0.5695 &0.3464 &0.3781 &0.3271 \\
\textit{w/o temporal} &0.3719 &0.4402 &0.2811 &0.3032 &0.2691 &0.4710 &0.5715 &0.3419 &0.3747 &0.3222 \\
\textit{w/o relaying} &0.3709 &0.4403 &0.2793 &0.3018 &0.2674 &0.4791 &0.5776 &0.3516 &0.3831 &0.3327 \\
\textit{w/o spatiotemporal} &0.3458 &0.4125 &0.2633 &0.2849 &0.2527 &0.4538 &0.5470 &0.3305 &0.3607 &0.3109 \\ \hline
STRealy+LoTNext &\textbf{0.3744} &\textbf{0.4455} &\textbf{0.2821} &\textbf{0.3051} &\textbf{0.2704} &\textbf{0.4856} &\textbf{0.5828} &\textbf{0.3612} &\textbf{0.3929} &\textbf{0.3423} \\
\textit{w/o spatial} &0.3652 &0.4337 &0.2772 &0.2993 &0.2665 &0.4733 &0.5690 &0.3509 &0.3821 &0.3325 \\
\textit{w/o temporal} &0.3685 &0.4372 &0.2789 &0.3011 &0.2675 &0.4746 &0.5700 &0.3489 &0.3800 &0.3294 \\
\textit{w/o relaying} &0.3742 &0.4451 &0.2817 &0.3045 &0.2695 &0.4829 &0.5786 &0.3573 &0.3885 &0.3377 \\
\textit{w/o spatiotemporal} &0.3530 &0.4199 &0.2684 &0.2901 &0.2587 &0.4581 &0.5534 &0.3350 &0.3660 &0.3165 \\ \hline
STRealy+Graph-Flashback &\textbf{0.3807} &\textbf{0.4525} &\textbf{0.2873} &\textbf{0.3101} &\textbf{0.2743} &\textbf{0.4863} &\textbf{0.5835} &\textbf{0.3555} &\textbf{0.3862} &\textbf{0.3381} \\
\textit{w/o spatial} &0.3737 &0.4432 &0.2827 &0.3052 &0.2709 &0.4784 &0.5763 &0.3526 &0.3844 &0.3333 \\
\textit{w/o temporal} &0.3766 &0.4453 &0.2845 &0.3068 &0.2723 &0.4764 &0.5773 &0.3477 &0.3805 &0.3279 \\
\textit{w/o relaying} &0.3684 &0.4368 &0.2787 &0.3009 &0.2672 &0.4770 &0.5741 &0.3503 &0.3819 &0.3305 \\
\textit{w/o spatiotemporal} &0.3521 &0.4173 &0.2711 &0.2925 &0.2578 &0.4589 &0.5548 &0.3346 &0.3647 &0.3154 \\ \hline
\end{tabular}
\vspace{-1em}
\end{table*}

\begin{table*}[!htp]\centering
\caption{Performance Comparison of Different STRelay Variants on Singapore and Moscow Datasets}
\vspace{-1em}
\label{tab:ablation_results_sin_mos}
\setlength{\tabcolsep}{0.5em}
\begin{tabular}{l|ccccc|ccccc} \hline
 & \multicolumn{5}{c|}{Singapore} & \multicolumn{5}{c}{Moscow} \\ \hline
Methods &Acc@5 &Acc@10 &NDCG@5 &NDCG@10 &MRR &Acc@5 &Acc@10 &NDCG@5 &NDCG@10 &MRR \\  \hline
STRelay+STGN &\textbf{0.2914} &\textbf{0.3748} &\textbf{0.2081} &\textbf{0.2347} &\textbf{0.2024} &\textbf{0.3841} &\textbf{0.4633} &\textbf{0.2819} &\textbf{0.3076} &\textbf{0.2669} \\
\textit{w/o spatial} &0.2832 &0.3597 &0.2039 &0.2286 &0.1980 &0.3667 &0.4361 &0.2675 &0.2901 &0.2517 \\
\textit{w/o temporal} &0.2793 &0.3550 &0.2013 &0.2258 &0.1958 &0.3756 &0.4443 &0.2743 &0.2966 &0.2574 \\
\textit{w/o relaying} &0.2881 &0.3659 &0.2079 &0.2330 &0.2017 &0.3697 &0.4420 &0.2713 &0.2948 &0.2562 \\
\textit{w/o spatiotemporal} &0.2736 &0.3538 &0.1950 &0.2209 &0.1905 &0.3475 &0.4157 &0.2533 &0.2755 &0.2392 \\ \hline
STRealy+Flashback &\textbf{0.3116} &\textbf{0.3978} &\textbf{0.2243} &\textbf{0.2523} &\textbf{0.2175} &\textbf{0.4092} &\textbf{0.4861} &\textbf{0.3001} &\textbf{0.3251} &\textbf{0.2816} \\
\textit{w/o spatial} &0.3075 &0.3902 &0.2223 &0.2491 &0.2155 &0.3941 &0.4688 &0.2894 &0.3137 &0.2723 \\
\textit{w/o temporal} &0.3106 &0.3939 &0.2238 &0.2508 &0.2167 &0.4044 &0.4764 &0.2977 &0.3212 &0.2794 \\
\textit{w/o relaying} &0.3092 &0.3942 &0.2235 &0.2511 &0.2173 &0.3965 &0.4707 &0.2914 &0.3156 &0.2740 \\
\textit{w/o spatiotemporal} &0.2995 &0.3815 &0.2159 &0.2425 &0.2092 &0.3780 &0.4410 &0.2802 &0.3007 &0.2633 \\ \hline
STRealy+SNPM &\textbf{0.3279} &\textbf{0.4172} &\textbf{0.2382} &\textbf{0.2663} &\textbf{0.2259} &\textbf{0.4167} &\textbf{0.4944} &\textbf{0.3047} &\textbf{0.3288} &\textbf{0.2877} \\
\textit{w/o spatial} &0.3169 &0.4040 &0.2261 &0.2543 &0.2185 &0.4101 &0.4827 &0.3034 &0.3270 &0.2851 \\
\textit{w/o temporal} &0.3199 &0.4115 &0.2279 &0.2575 &0.2206 &0.4137 &0.4877 &0.3043 &0.3284 &0.2853 \\
\textit{w/o relaying} &0.3162 &0.4015 &0.2259 &0.2536 &0.2182 &0.4140 &0.4899 &0.3039 &0.3286 &0.2859 \\
\textit{w/o spatiotemporal} &0.3138 &0.3993 &0.2268 &0.2520 &0.2164 &0.3913 &0.4562 &0.2906 &0.3118 &0.2738 \\ \hline
STRealy+LoTNext &\textbf{0.3312} &\textbf{0.4188} &\textbf{0.2402} &\textbf{0.2686} &\textbf{0.2323} &\textbf{0.4192} &\textbf{0.4949} &\textbf{0.3092} &\textbf{0.3338} &\textbf{0.2905} \\
\textit{w/o spatial} &0.3274 &0.4170 &0.2379 &0.2670 &0.2307 &0.4163 &0.4897 &0.3081 &0.3320 &0.2894 \\
\textit{w/o temporal} &0.3264 &0.4141 &0.2378 &0.2662 &0.2305 &0.4183 &0.4948 &0.3080 &0.3323 &0.2881 \\
\textit{w/o relaying} &0.3277 &0.4163 &0.2389 &0.2676 &0.2318 &0.4182 &0.4927 &0.3090 &0.3337 &0.2903 \\
\textit{w/o spatiotemporal} &0.3241 &0.4093 &0.2342 &0.2618 &0.2260 &0.4025 &0.4719 &0.2990 &0.3216 &0.2814 \\ \hline
STRealy+Graph-Flashback &\textbf{0.3348} &\textbf{0.4251} &\textbf{0.2433} &\textbf{0.2728} &\textbf{0.2338} &0.4235 &\textbf{0.4993} &\textbf{0.3113} &\textbf{0.3360} &\textbf{0.2915} \\
\textit{w/o spatial} &0.3220 &0.4060 &0.2322 &0.2594 &0.2240 &0.4180 &0.4934 &0.3086 &0.3331 &0.2898 \\
\textit{w/o temporal} &0.3214 &0.4052 &0.2319 &0.2590 &0.2237 &\textbf{0.4241} &0.4986 &0.3105 &0.3349 &0.2903 \\
\textit{w/o relaying} &0.3313 &0.4219 &0.2392 &0.2686 &0.2324 &0.4117 &0.4827 &0.3025 &0.3256 &0.2832 \\
\textit{w/o spatiotemporal} &0.3207 &0.4041 &0.2314 &0.2584 &0.2233 &0.3986 &0.4645 &0.2967 &0.3197 &0.2765 \\ \hline
\end{tabular}
\vspace{-1em}
\end{table*}

\subsubsection{Experiment Settings} 
We developed our model using the PyTorch framework and conducted experiments on the following hardware platform (CPU: Intel(R) Xeon(R) Gold 5320, GPU: NVIDIA GeForce RTX 3090). We evaluate STRealy and the baselines in the location prediction task. The implementations of all baselines and base models are either provided by the original authors or based on the original papers.

\subsection{Overall Performance Comparison}
Table \ref{tab:all_results_ist_tky} and Table \ref{tab:all_results_sin_mos} show the overall performance on all four datasets, and for each dataset and each metric, boldface indicates the best overall result, while underline highlights the strongest baseline method. We observe that among all STRelay enhanced methods, the STRelay+Graph-Flashback achieves the best performance in most cases, except on NDCG and MRR on the Tokyo dataset, where the STRelay+LoTNext achieves slightly better performance than STRelay+Graph-Flashback. More importantly, compared with the best-performing baselines, STRelay+Graph-Flashback yields average relative improvements of 6.72\%, 5.92\%, 3.74\%, and 4.45\% on the Istanbul, Tokyo, Singapore, and Moscow datasets, respectively, demonstrating its effectiveness. Moreover, DeepMove, STAN and GETNext consider future temporal information but still yield suboptimal performance compared to STRelay enhanced baselines.


Notably, we observe that our STRelay framework consistently improves the location prediction performance of the corresponding base prediction methods STGN, Flashback, Graph-Flashback, SNPM, and LoTNext, on all datasets, which demonstrates the universal effectiveness of the STRelay framework. STRelay boosts the performance of the above five baselines with an average improvement of 2.49\%-11.30\% across different datasets in the next location prediction task. 

In the following, we conduct a systematic ablation study below to verify the usefulness of different components of our STRelay framework.

\subsection{Ablation Study}
We consider four variants to quantify their impact on location prediction performance and demonstrate their utility. 
\begin{itemize}
    \item \textbf{\textit{w/o spatial}} is a variant that removes future spatial context. It is also equivalent to only considering future temporal contexts for location prediction.
    \item \textbf{\textit{w/o temporal}} is a variant that removes future temporal context. It is also equivalent to only considering future spatial contexts for location prediction.
    \item \textbf{\textit{w/o relaying}} is a variant that models future temporal and spatial contexts not in a relaying manner, but in a parallel manner. Specifically, we decoupled spatial and temporal context modeling (i.e., removing the sequential dependency where spatial modeling conditions on temporal modeling).
    \item \textbf{\textit{w/o spatiotemporal}} is a variant that removes both future spatial and temporal contexts, reducing STRelay to the plain base model.
\end{itemize}

The ablation results are reported in Tables \ref{tab:ablation_results_ist_tky} and \ref{tab:ablation_results_sin_mos}. First, we observe that the complete STRelay framework consistently outperforms the spatial-ablated variant (\textit{w/o spatial}) by 2.65\% on average across five base models and four datasets, highlighting the essential contributions of modeling future spatial context. Meanwhile, we observe that the \textit{w/o temporal} variant (which retains only future spatial contexts) outperforms the \textit{w/o spatiotemporal} variant by 3.89\% on average across all five base models and four datasets, further validating the usefulness of modeling future spatial context.

Similarly, compared to the complete STRelay framework, the temporal-ablated variant (\textit{w/o temporal}) that removes future temporal contexts, leads to an average performance drop of 2.31\%, confirming the critical importance of future temporal context. The usefulness of modeling future temporal context is also validated by the \textit{w/o spatial} variant (which retains only future temporal context) outperforms the \textit{w/o spatiotemporal} variant by 3.46\% on average across all five base models and four datasets.

Moreover, we notice that the complete STRelay framework outperforms the \textit{w/o relaying} variant by 0.47\%-2.90\% performance improvement across five base models and datasets, which demonstrates that our learning in a relaying manner indeed boosts performance by flexibly capturing the inherent relationship between temporal and spatial contexts.

Finally, the complete STRelay framework outperforms the context-free variant (\textit{w/o spatiotemporal}) by 6.28\% on average across all five base models and four datasets. This validates our core motivation, i.e., future spatiotemporal contexts provide highly informative signals for next-location prediction, and demonstrates the superiority of the proposed STRelay framework.


\begin{figure}[]
\centering
        {
        \includegraphics[width=0.65\linewidth]{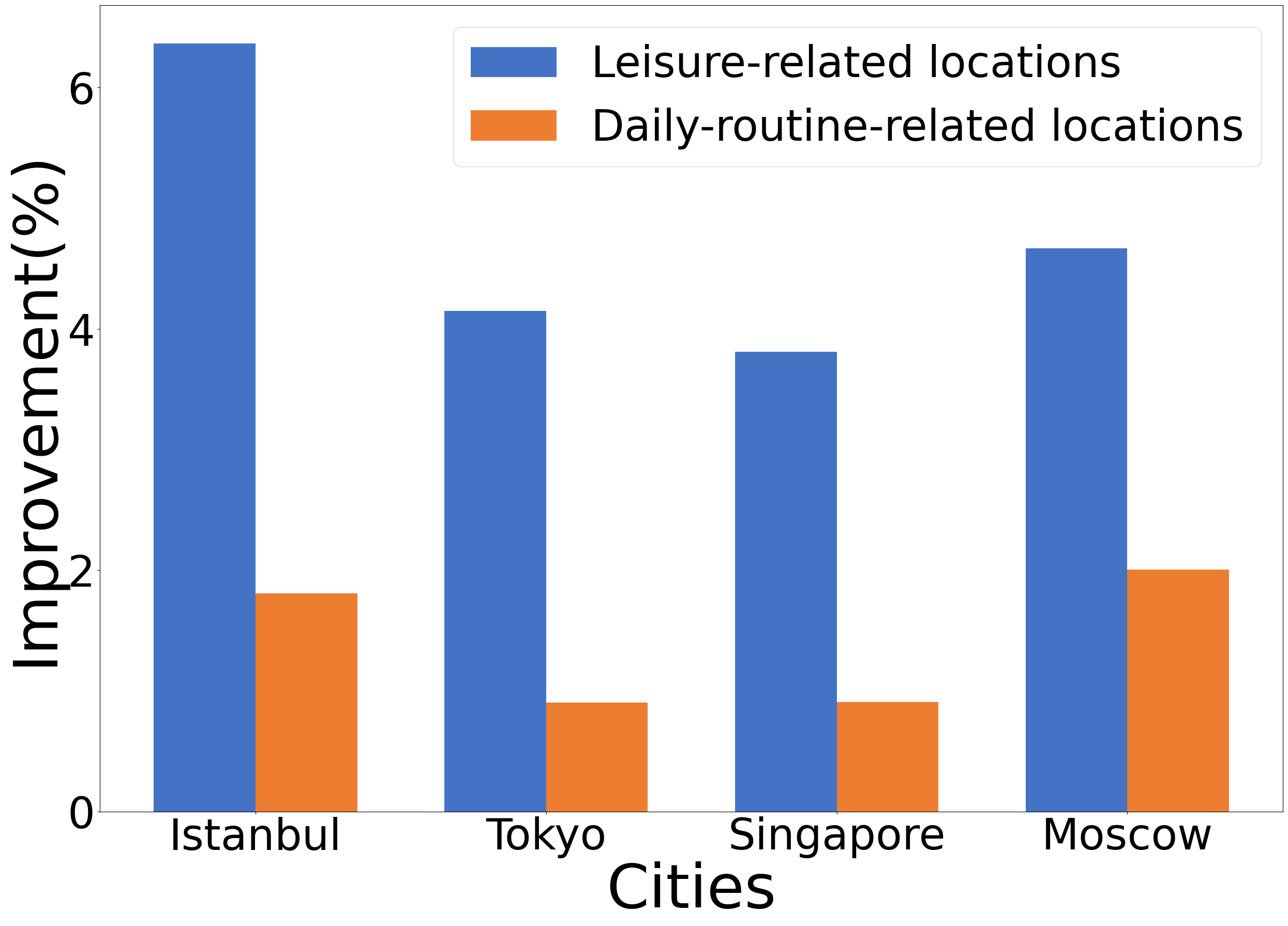} }
        
    \caption{Performance improvement on different location types}
    \label{fig:entertainment}
\end{figure}

\begin{figure}[]
\centering
    {
        \includegraphics[width=0.65\linewidth]{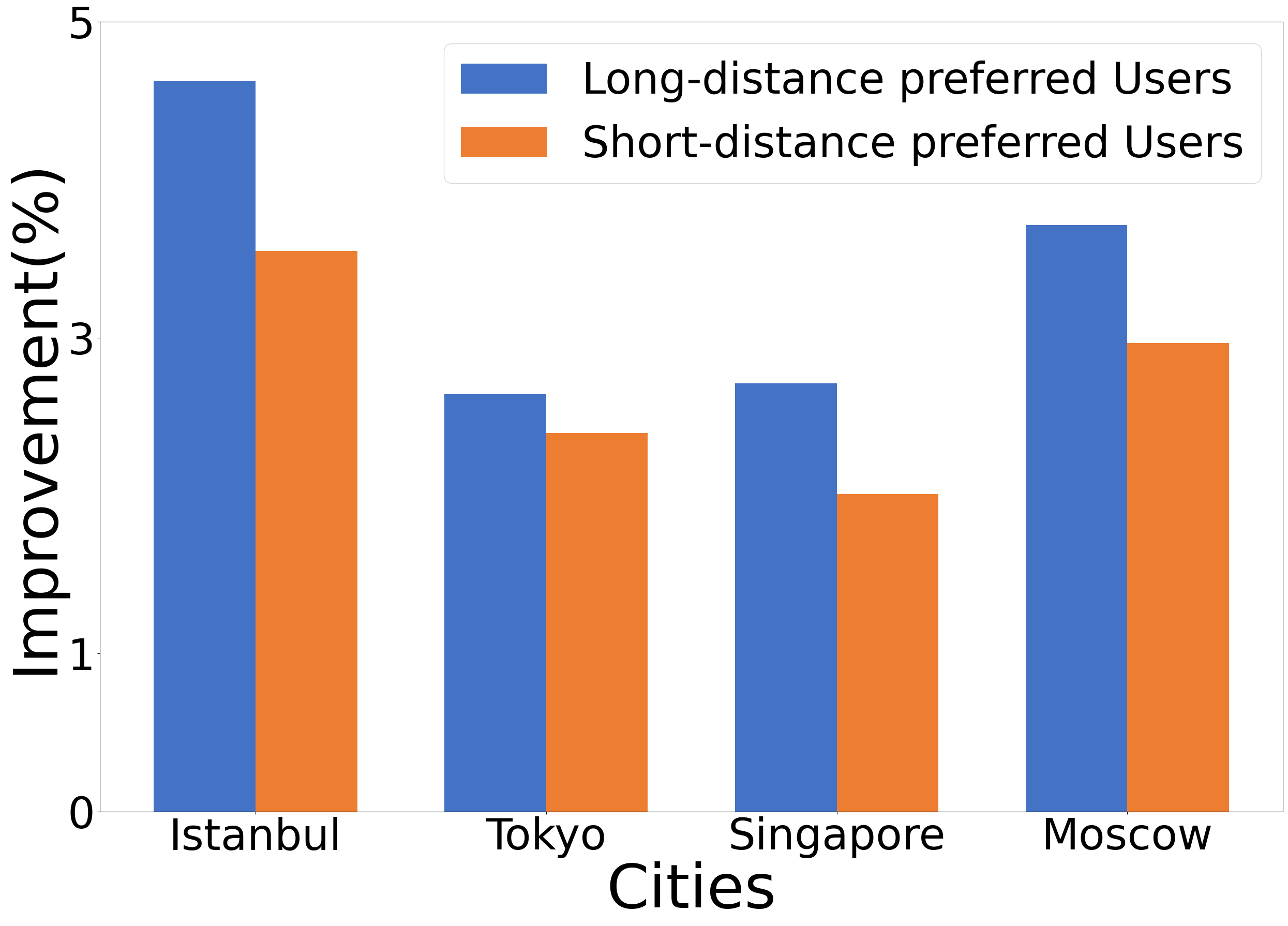}}
        
    \caption{Performance improvement on different user groups}
    \label{fig:improve_user}
\end{figure}

\subsection{Performance Improvement Analysis on Location Types and User Groups}
\label{motivation_experiment}
In this section, we investigate the performance improvement across different location types and user groups using STRelay with Graph-Flashback (the best-performing base model).

First, to study the performance improvement across different location types, we classify locations into corresponding activity categories following previous work \cite{yang2019revisiting}. Specifically, we adopt a binary classification scheme consisting of: leisure-related locations (e.g., Outdoors and Entertainment) and daily-routine-related locations (e.g., College and Residence). Figure \ref{fig:entertainment} shows the results of the percentage of performance improvement of $\text{Acc}$@10 in predicting these two types of locations on the four datasets. We observe that the performance improvements stemming from leisure-related locations are consistently and significantly higher than those from daily-routine-related locations by 2.66\%-4.57\%, which demonstrates that modeling future spatiotemporal contexts is particularly beneficial for entertainment-related activities, which are more uncertain than daily-routine-related activities.

Second, we categorize users into two distinct groups based on their radius of gyration \cite{movesim, deng2025revisiting}, which is calculated as the root mean squared distance of all locations in a trajectory from the geographical center of the trajectory, and obtain two groups: long-distance preferred users (top 50\%) and short-distance preferred users (bottom 50\%). We report the performance improvements on these two user groups. As shown in Figure \ref{fig:improve_user}, both user groups exhibit performance improvements in $\text{Acc}$@10 across all four datasets, with long-distance preferred users consistently demonstrating larger improvement (0.25\%-1.07\% higher than short-distance preferred users). These results indicate that modeling future spatiotemporal contexts is more effective in capturing the mobility patterns of users who prefer traveling longer distances, which often implies higher uncertainty for next locations.

\begin{figure}[ht]
    \subfigure{
        \includegraphics[width=0.47\linewidth]{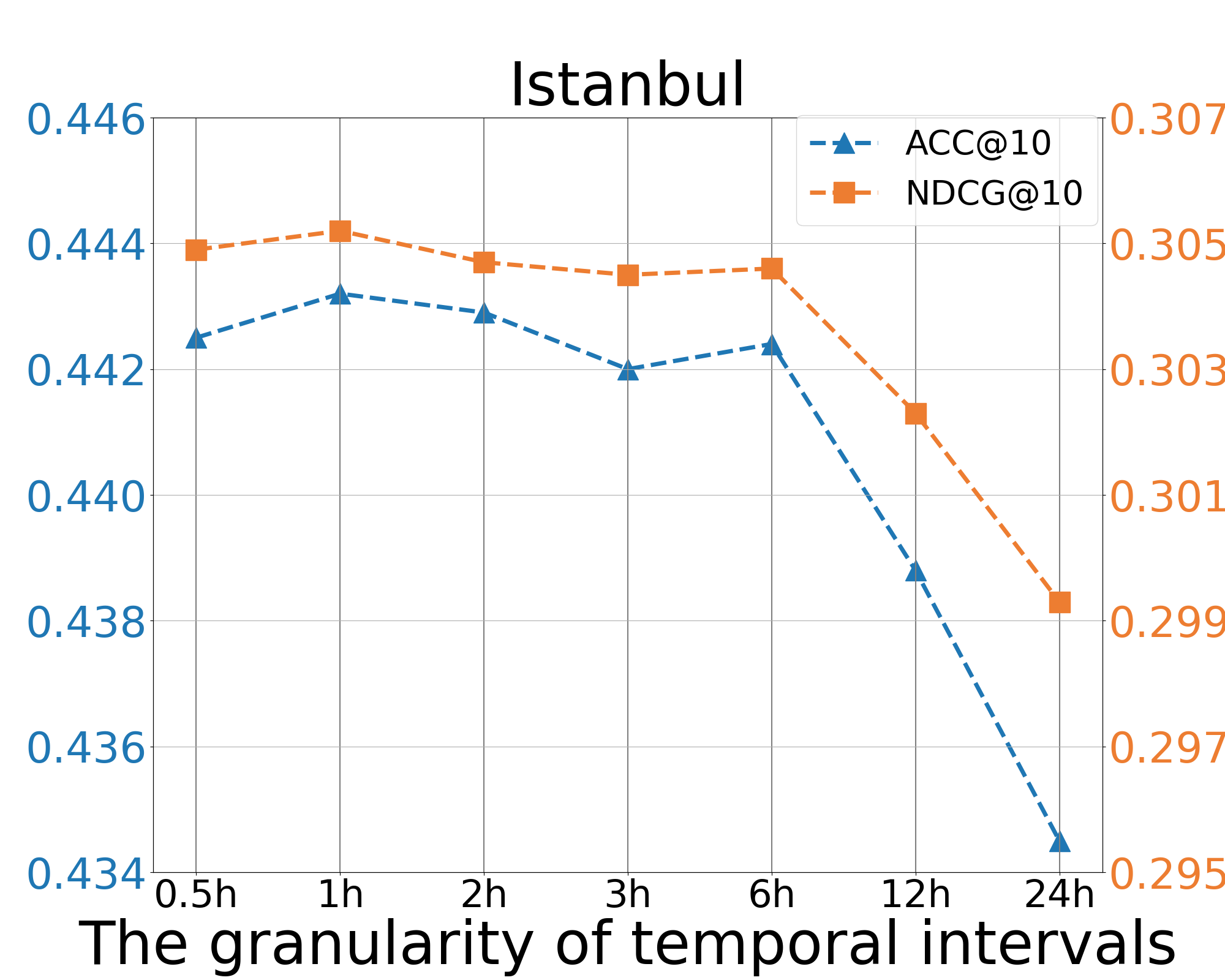} 
        \label{fig:ist_t}}
    \subfigure{
        \includegraphics[width=0.47\linewidth]{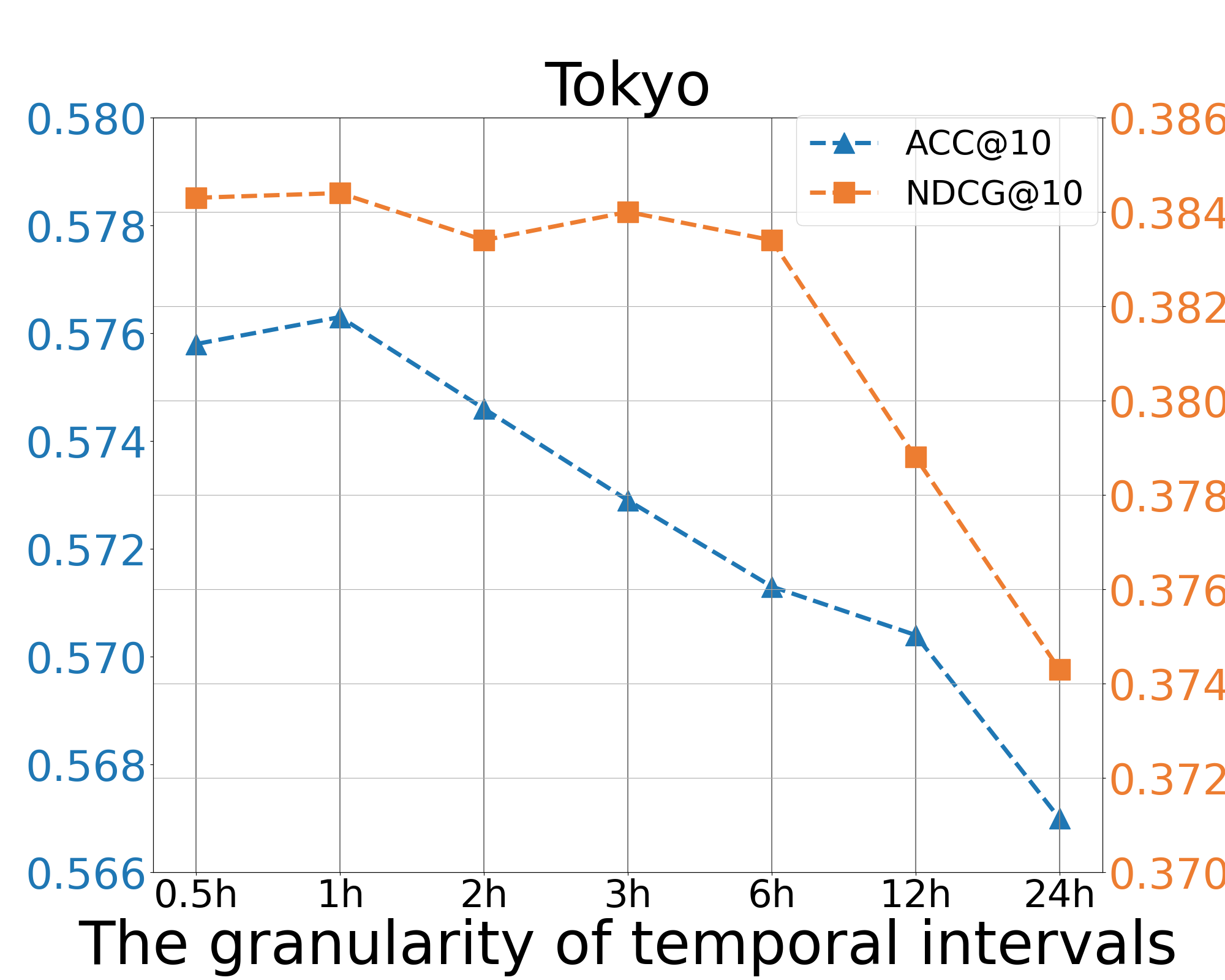} 
        \label{fig:tky_t}}
    \subfigure{
        \includegraphics[width=0.47\linewidth]{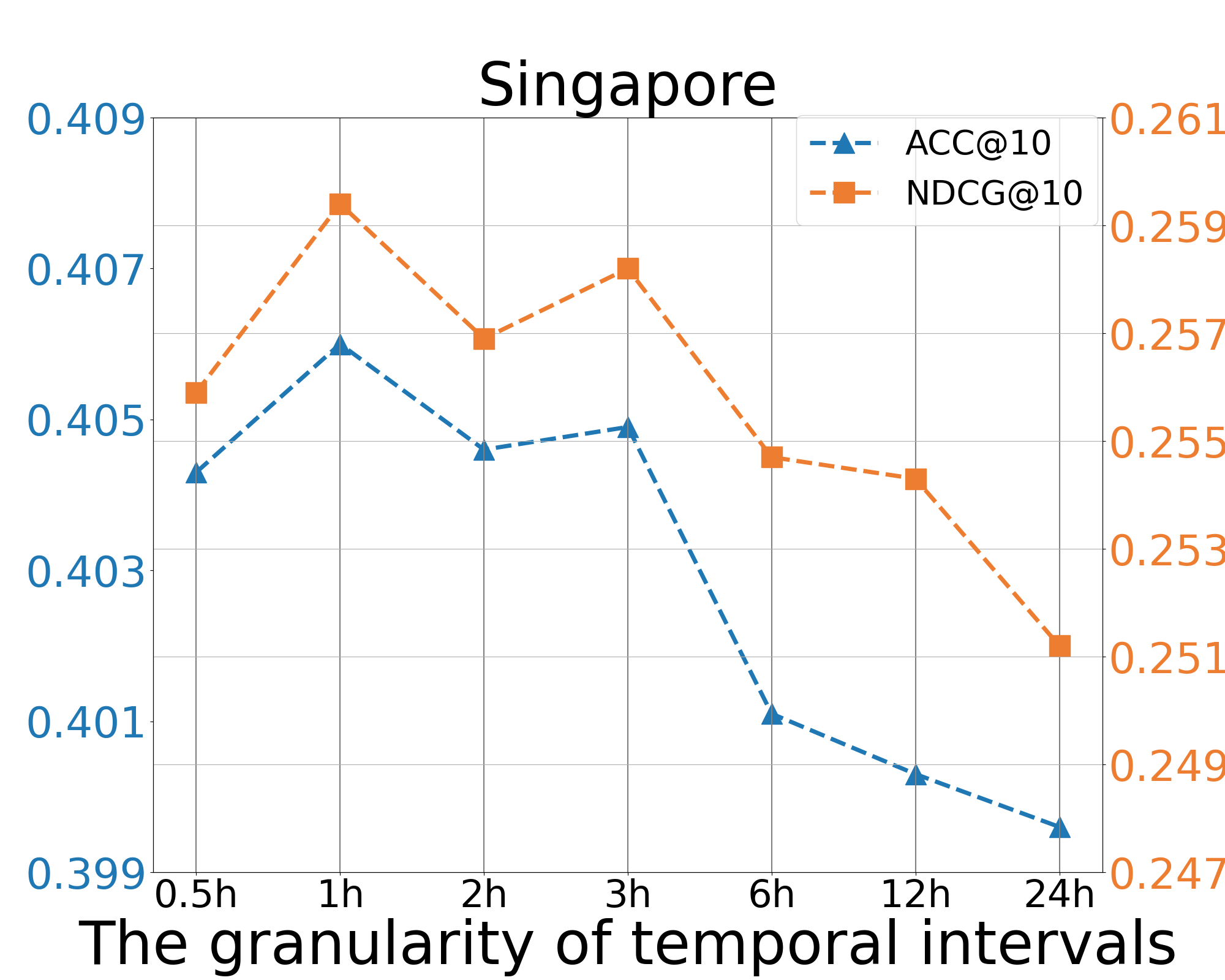} 
        \label{fig:sgp_t}}
    \subfigure{
        \includegraphics[width=0.47\linewidth]{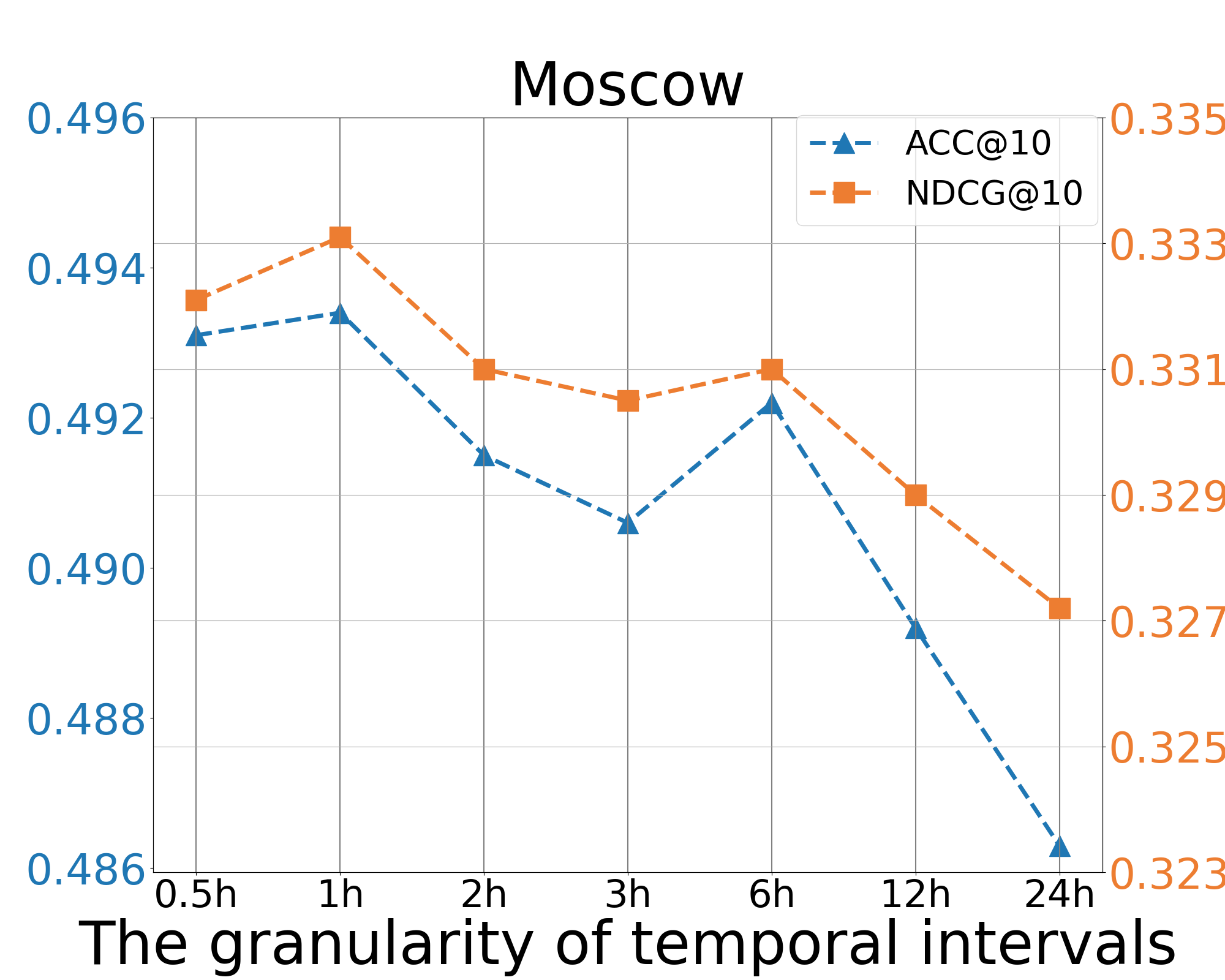} 
        \label{fig:moscow_t}}
    \caption{Impact of the  granularity of temporal intervals}
    \label{fig:parameter_time}
\end{figure}

\begin{figure}[ht]
    \subfigure{
        \includegraphics[width=0.47\linewidth]{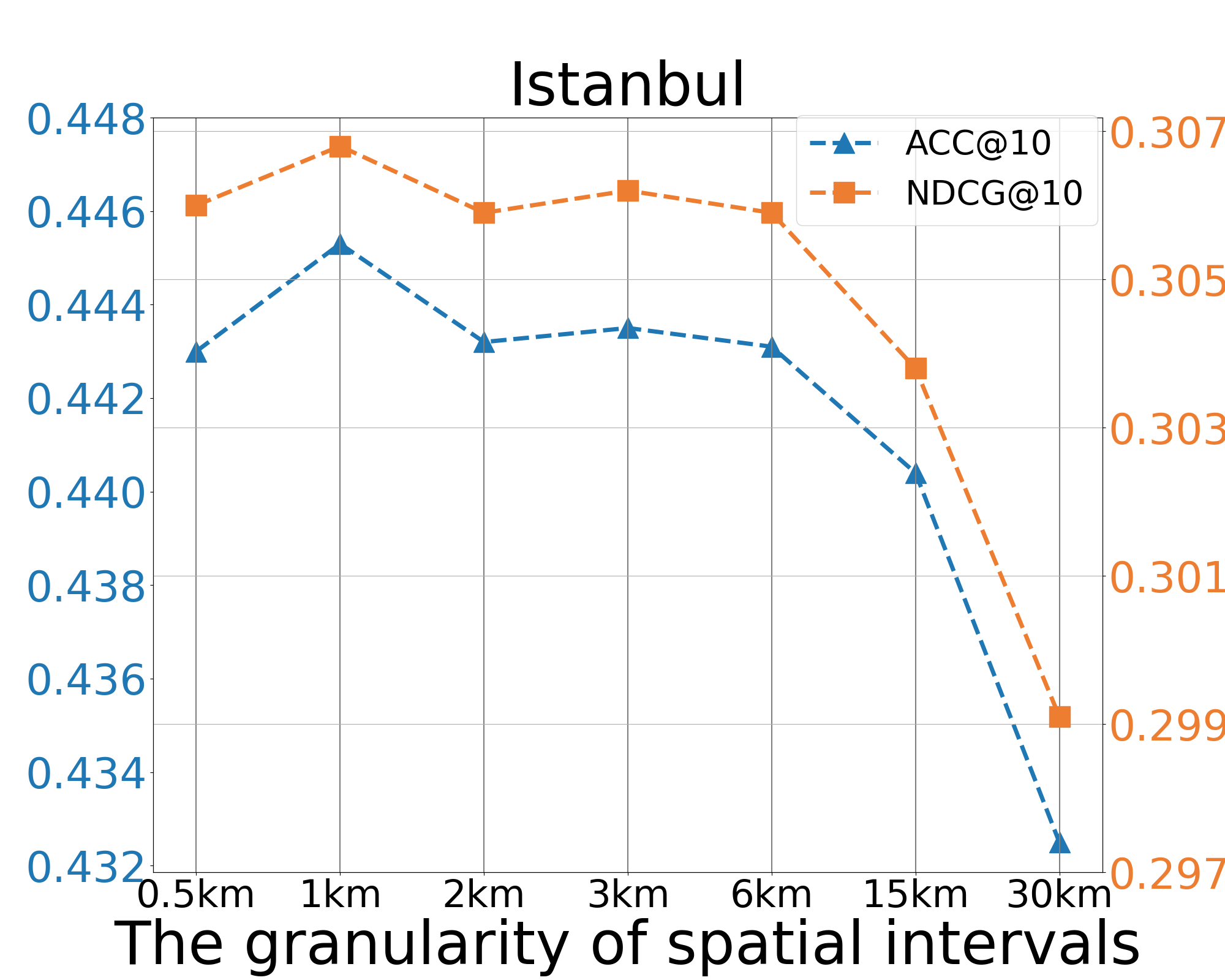} 
        \label{fig:ist_d}}
    \subfigure{
        \includegraphics[width=0.47\linewidth]{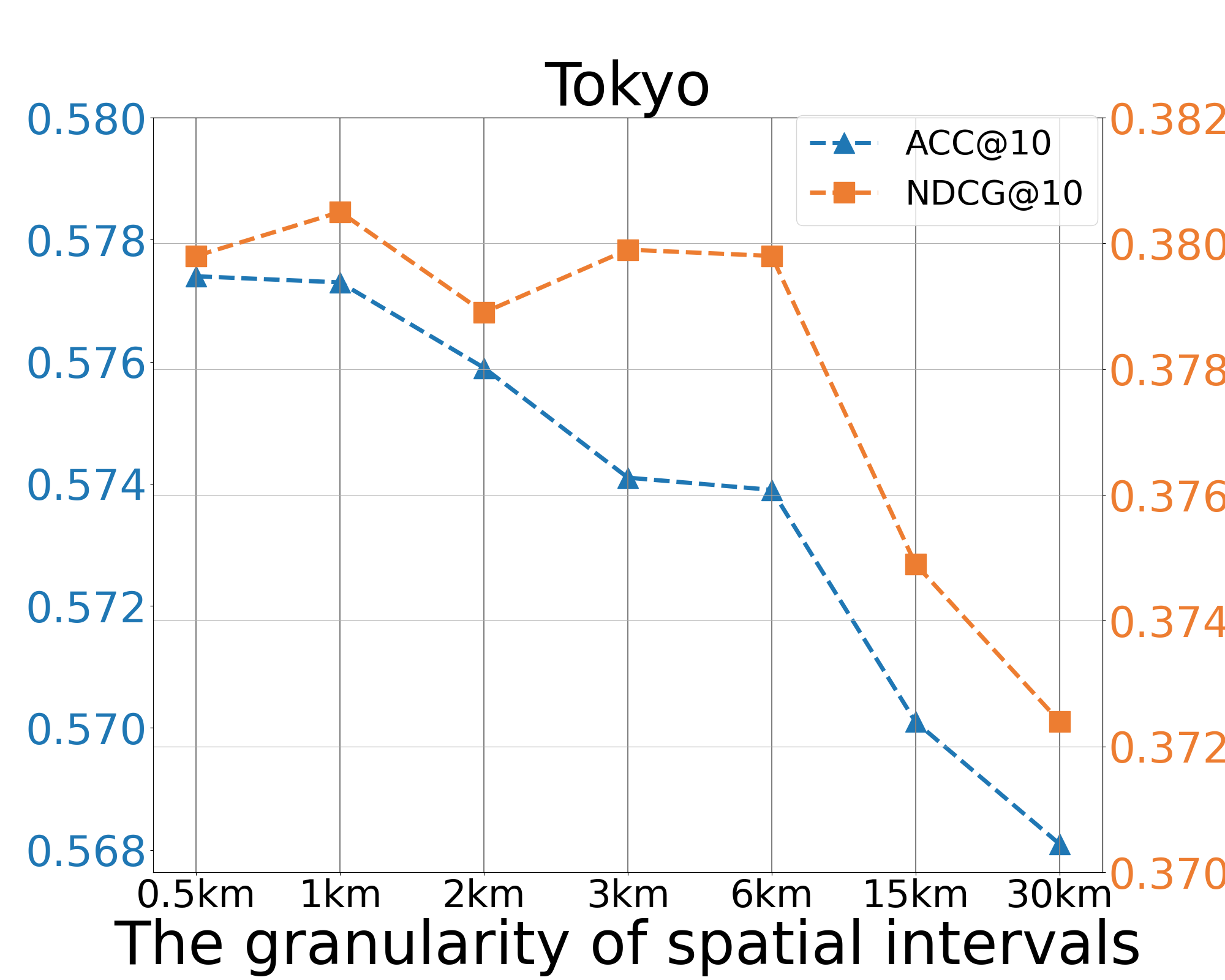} 
        \label{fig:tky_d}}
    \subfigure{
        \includegraphics[width=0.47\linewidth]{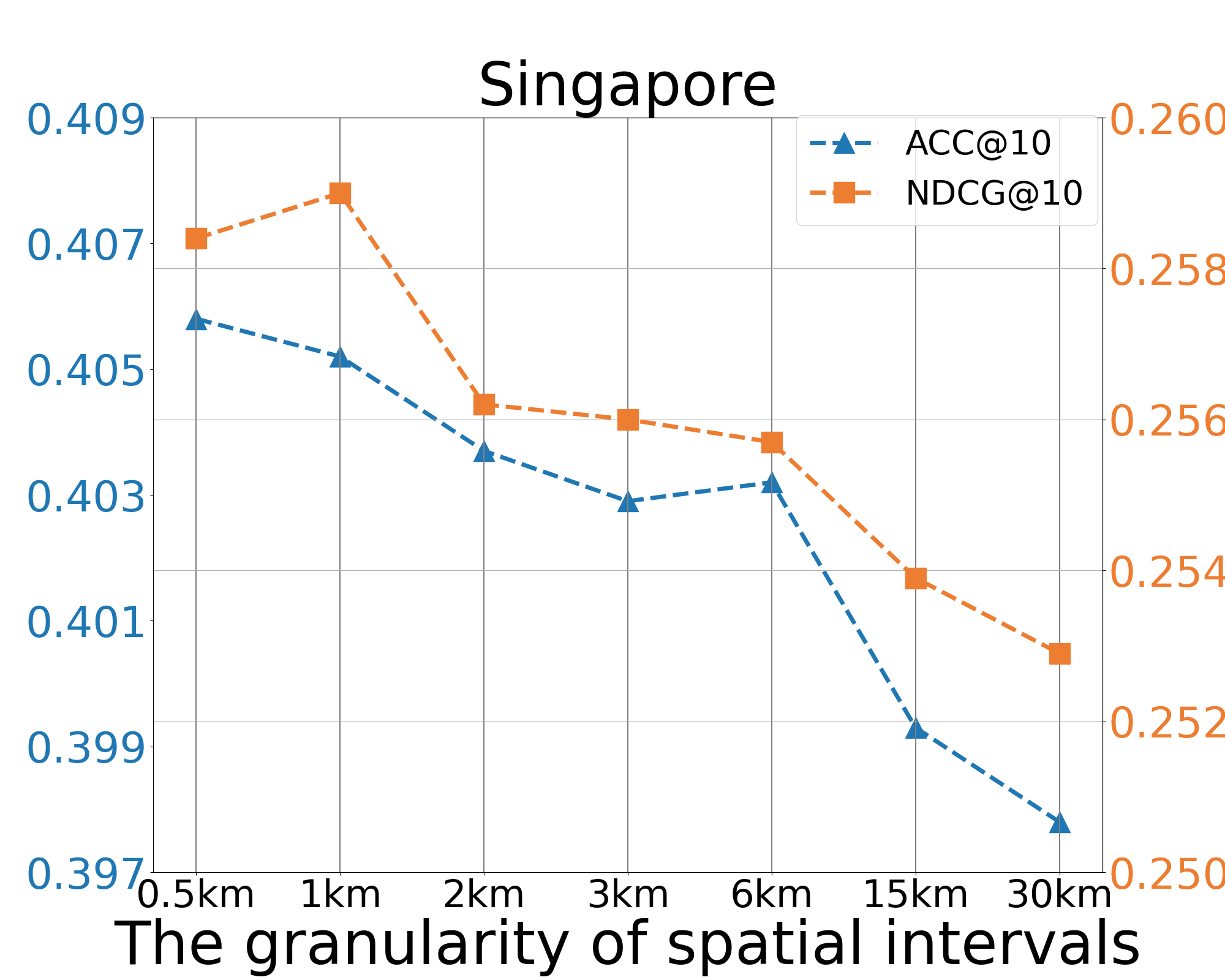} 
        \label{fig:sgp_d}}
    \subfigure{
        \includegraphics[width=0.47\linewidth]{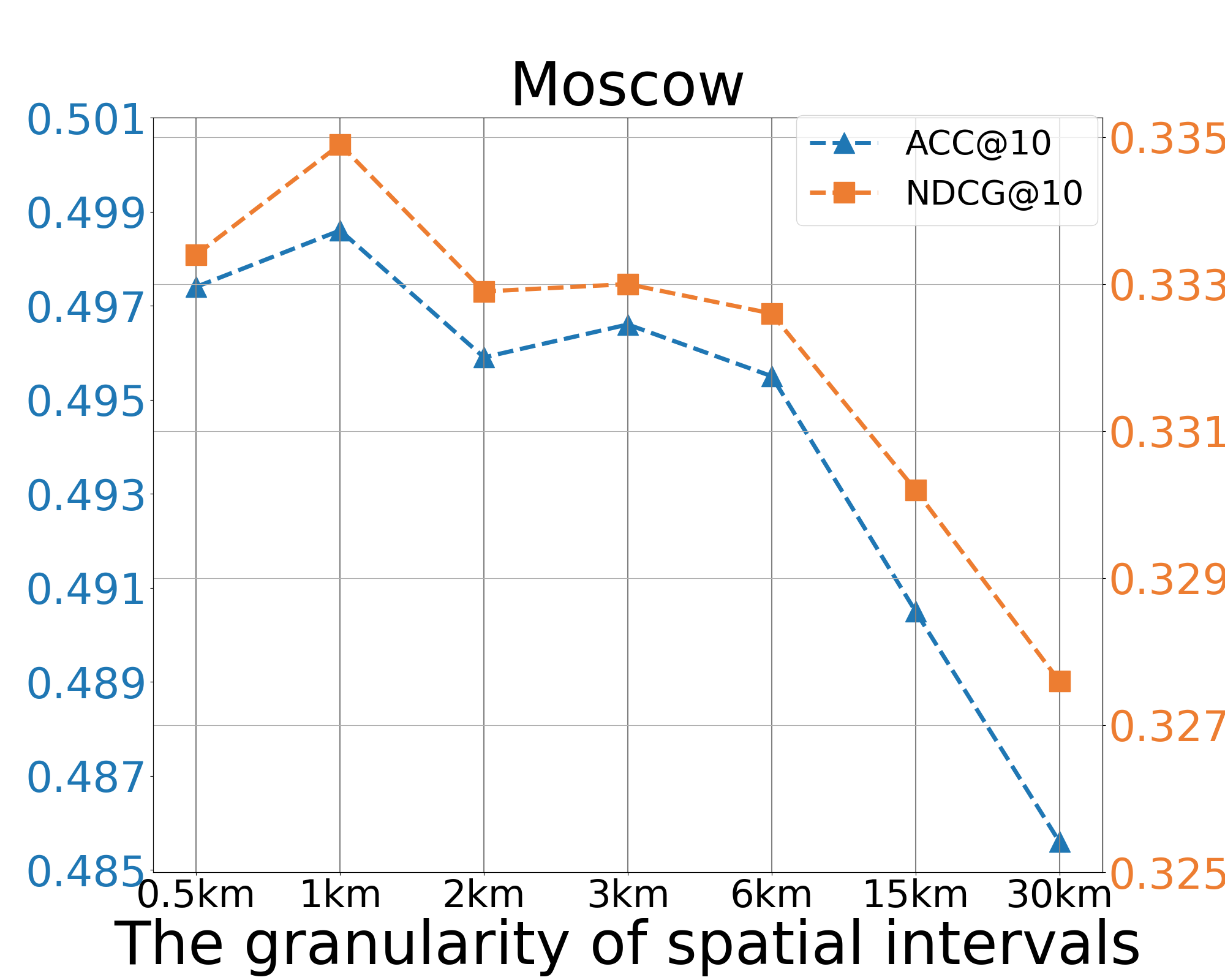} 
        \label{fig:msk_d}}   
    \caption{Impact of the granularity of spatial intervals}
    \label{fig:parameter_distance}
\end{figure}

\subsection{Impact of Spatiotemporal Context Granularity}
In this section, we use STRelay with Graph-Flashback as the base model to investigate the influence of the granularity of future temporal and spatial intervals. 

Firstly, to investigate the impact of temporal context granularity, we train the model with future time intervals ranging from 0.5 to 24 hours. We report the results of Acc@10 and NDCG@10 on four datasets as shown in Figure \ref{fig:parameter_time}. We observe that both metrics exhibit a clear declining trend as the temporal granularity becomes coarser (i.e., longer intervals lead to poorer performance). Notably, the 1-hour granularity consistently yields the highest Acc@10 and NDCG@10 on all datasets. Based on these observations, we adopt 1 hour as the optimal temporal granularity in our experiment settings.

Secondly, to investigate the impact of spatial context granularity, we train the model with spatial intervals ranging from 0.5 km to 30 km. The corresponding results of Acc@10 and NDCG@10 are presented in Figure \ref{fig:parameter_distance}. Both metrics follow the same trend as observed for temporal granularity: the performance degrades as the spatial interval becomes coarser.  
Although the 0.5 km setting performs slightly better on the Singapore and Tokyo datasets, the 1 km granularity delivers the highest scores on the Istanbul and Moscow datasets, while achieving the best NDCG@10 across all four datasets. Therefore, we select 1 km as the optimal spatial granularity in our experiment settings.


\section{Conclusion}
In this paper, we propose STRelay, a universal framework designed to explicitly model the future spatiotemporal contexts in a relaying manner; it can be flexibly integrated with different location prediction models to boost the prediction performance. Experimental results show that STRelay consistently boosts the performance of state-of-the-art baselines with an average improvement of 2.49\%-11.30\% across four datasets, demonstrating the universal effectiveness of STRelay. Moreover, we find that the future spatiotemporal contexts are particularly helpful for entertainment-related locations and also for user groups who prefer traveling longer distances. The performance gain on such non-daily-routine activities, which often suffer from higher uncertainty, is indeed complementary to the base location prediction models that excel at modeling regular daily routine patterns. 

In the future, we plan to learn future spatiotemporal contexts using LLM to get better generalizability across cities.



\bibliographystyle{IEEEtran}
\bibliography{manuscript}

\end{document}